\documentclass[10pt,journal,compsoc]{IEEEtran}
\usepackage{amsmath,amsfonts}
\usepackage{algorithm}
\usepackage{array}
\usepackage[caption=false,font=normalsize,labelfont=sf,textfont=sf]{subfig}
\usepackage{textcomp}
\usepackage{stfloats}

\usepackage{verbatim}
\usepackage{graphicx}
\usepackage[dvipsnames]{xcolor}

\usepackage{amssymb}
\usepackage{booktabs}
\usepackage{multirow}
\usepackage{bm}
\usepackage{enumitem}
\usepackage{colortbl}
\usepackage{mathtools}
\usepackage{xcolor}

\usepackage{algpseudocode}
\usepackage{tabularx}
\usepackage{url}
\usepackage{color,soul}
\usepackage{epsfig}

\usepackage{multirow}
\usepackage{makecell}
\usepackage{pifont}
\usepackage[switch]{lineno}
\usepackage[misc]{ifsym}

\usepackage[colorlinks,citecolor=blue,urlcolor=blue,linkcolor=blue]{hyperref}
\usepackage{cleveref}
\usepackage{url}
\usepackage[dvipsnames]{xcolor}
\usepackage{subcaption}
\definecolor{mygray}{HTML}{E9F1F6}
\newcommand{\cmark}{\ding{51}} 
\newcommand{\xmark}{\ding{55}} 
\usepackage{tabularx}

\definecolor{shapecolor}{rgb}{0.0,0.5,0.0}

\usepackage{newtxtext,newtxmath}

\usepackage{amsmath,amsfonts,bm}

\def\eqref#1{equation~\ref{#1}}

\def\1{\bm{1}}

\def\vx{{\bm{x}}}

\DeclareMathAlphabet{\mathsfit}{\encodingdefault}{\sfdefault}{m}{sl}
\SetMathAlphabet{\mathsfit}{bold}{\encodingdefault}{\sfdefault}{bx}{n}

\def\gC{{\mathcal{C}}}
\def\gD{{\mathcal{D}}}

\def\gF{{\mathcal{F}}}

\def\gL{{\mathcal{L}}}

\usepackage{siunitx}
\title{Continuous Knowledge-Preserving Decomposition with Adaptive Layer Selection for Few-Shot Class-Incremental Learning}

\author{
Xiaojie Li, \quad
Jianlong Wu,~\IEEEmembership{Member,~IEEE,} \quad
Yue Yu,~\IEEEmembership{Member,~IEEE,} \\
Liqiang Nie,~\IEEEmembership{Senior Member,~IEEE,} \quad
Min Zhang

\IEEEcompsocitemizethanks{
\IEEEcompsocthanksitem Xiaojie Li, Jianlong Wu, Liqiang Nie, and Min Zhang are with the School of Computer Science and Technology, Harbin Institute of Technology (Shenzhen), China. E-mail: xiaojieli0903@gmail.com, wujianlong@hit.edu.cn, nieliqiang@gmail.com, zhangmin2021@hit.edu.cn.
\IEEEcompsocthanksitem Xiaojie Li, Jianlong Wu, and Yue Yu are with Pengcheng Laboratory, Shenzhen, China. E-mail: yuy@pcl.ac.cn.
}}

\begin{document}
\IEEEtitleabstractindextext{
\begin{abstract}
Few-Shot Class-Incremental Learning (FSCIL) faces a critical challenge: balancing the retention of prior knowledge with the acquisition of new classes. Existing methods either freeze the backbone to prevent catastrophic forgetting, sacrificing plasticity, or add new modules, incurring high costs. These approaches treat pretrained models as black boxes, overlooking two key opportunities to exploit their internal capacity: reusing redundant representational space within layers and selectively adapting layers based on their sensitivity to forgetting.
We propose CKPD-FSCIL, a unified framework that unlocks the underutilized capacity of pretrained weights, achieving a superior stability-plasticity balance with zero inference overhead. Our design integrates two continuously adapting mechanisms:
At the weight level, a Continuous Knowledge-Preserving Decomposition mechanism uses feature covariance to split each weight matrix into a frozen subspace that safeguards prior knowledge and a learnable, redundant subspace for new tasks.
At the layer level, a Continuous Adaptive Layer Selection mechanism leverages an Adapter Sensitivity Ratio to automatically select layers with the highest redundant capacity and lowest forgetting risk for adaptation.
By targeting only safe, high-potential subspaces and layers, CKPD-FSCIL enables efficient adaptation. After each session, the learned adapters are merged back into the original weights, ensuring zero additional parameters or FLOPs during inference. Extensive experiments on multiple FSCIL benchmarks demonstrate that our method consistently outperforms state-of-the-art approaches in both adaptability and knowledge retention. The code is available at \url{https://github.com/xiaojieli0903/CKPD-FSCIL}.
\end{abstract}
}
    
\maketitle
\IEEEdisplaynontitleabstractindextext
\IEEEpeerreviewmaketitle
\section{Introduction}

\begin{figure*}
  \centering
  \includegraphics[width=0.95\linewidth]{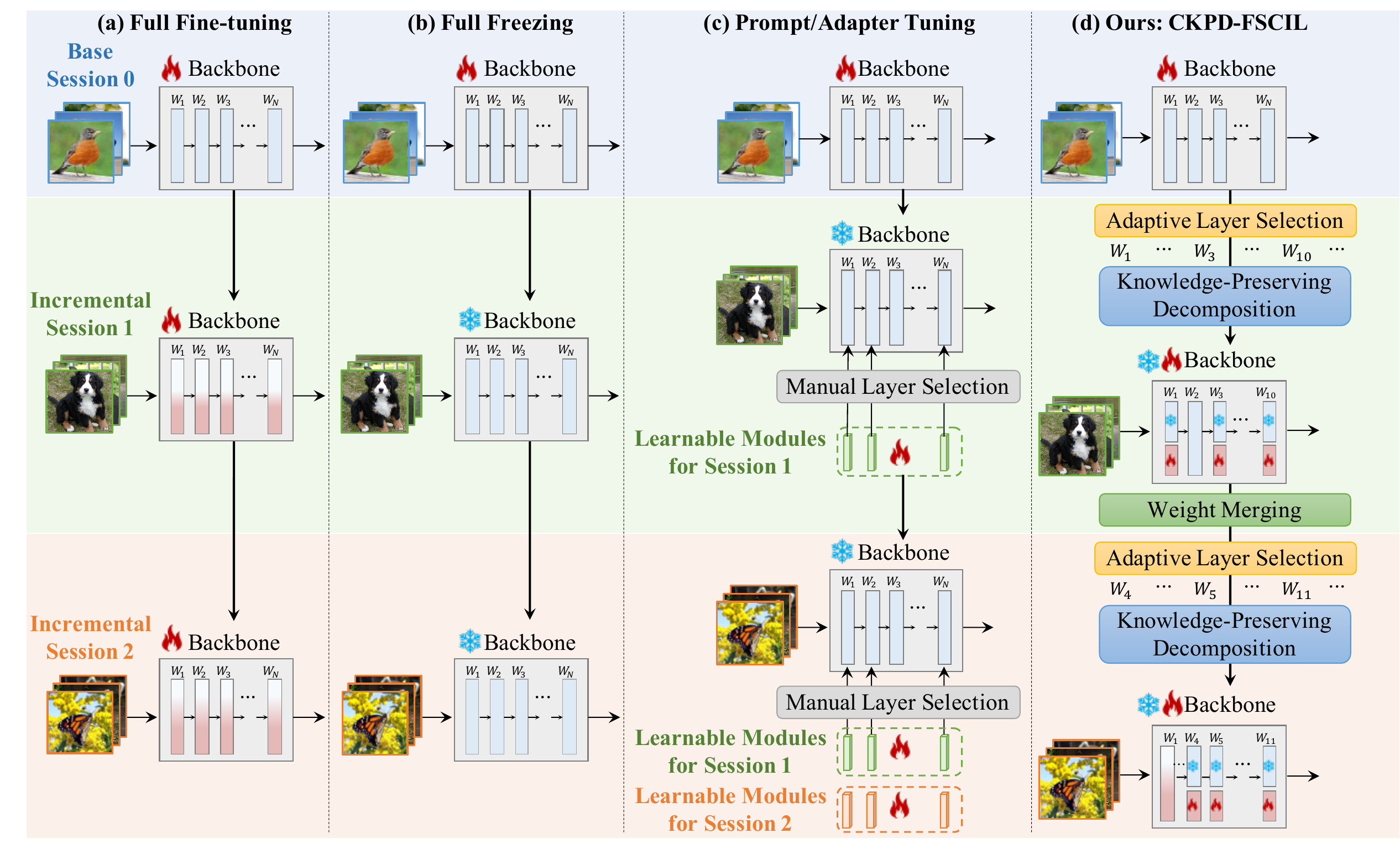}
\caption{
\textbf{Comparison of Parameter Adaptation Paradigms in FSCIL.} 
(a) \textbf{Full Fine-tuning}: Updates all backbone weights ($W_1, \dots, W_N$) in every session, offering high plasticity but suffering from overfitting and catastrophic forgetting. 
(b) \textbf{Full Freezing}: Keeps all backbone weights fixed after the base session, preserving prior knowledge but limiting the capacity for adaptation.
(c) \textbf{Prompt/Adapter Tuning}: Adds session-specific modules to a frozen backbone, enabling incremental adaptation but resulting in cumulative growth in parameters and inference cost.
(d) \textbf{Our CKPD-FSCIL}: Introduces a \textbf{select-decompose-train-merge} cycle. It intelligently identifies which layers to adapt via \textbf{Adaptive Layer Selection} and repurposes their internal weights by separating them into frozen knowledge-sensitive and learnable redundant components via \textbf{Knowledge-Preserving Decomposition}, enabling targeted adaptation without architectural growth.
}
\label{fig:motivation}
\vspace{-4mm}
\end{figure*}

Few-Shot Class-Incremental Learning (FSCIL)~\cite{tao2020few} aims to build systems that can incrementally learn novel classes from only a limited labeled samples, while preserving previously acquired knowledge~\cite{wang2024jarvis,lioptimus,zheng2025lifelong,li2024vision}. This setting is crucial for real-world applications like personalized recommendation, robotic perception in dynamic environments, and rare-condition medical diagnostics.
However, FSCIL poses two fundamental challenges: \emph{catastrophic forgetting}, where the acquisition of new knowledge disrupts or overwrites prior learning~\cite{mccloskey1989catastrophic, goodfellow2013empirical}; and \emph{overfitting}, due to the scarcity of labeled data for new classes~\cite{snell2017prototypical, sung2018learning}. Addressing the trade-off between \emph{stability} (preserving existing knowledge) and \emph{plasticity} (adapting to novel tasks)~\cite{mermillod2013stability} is central to continual learning.

Existing approaches typically follow three parameter adaptation paradigms to balance stability and plasticity, as shown in Fig.~\ref{fig:motivation}:
\textbf{(a) Full Fine-tuning} updates all backbone weights in each session, offering strong adaptability but often causing overfitting and forgetting~\cite{rebuffi2017icarl, hou2019learning, kirkpatrick2017overcoming, shin2017continual}.
\textbf{(b) Full Freezing} keeps the backbone fixed after the base session, effectively preserving prior knowledge but limiting the capacity to learn new tasks~\cite{zhang2021few, yang2023neural, li2024mamba}.
\textbf{(c) Prompt/Adapter Tuning} introduces lightweight, session-specific modules into a frozen backbone~\cite{park2024pre, ran2024brain, goswami2024calibrating, d2023multimodal, wang2024knowledge, liu2024few, tian2024pl}, offering a trade-off between adaptation and forgetting, but leading to growing inference and memory costs over time.

These approaches share a common limitation: they treat pretrained weights as a \textbf{black box}: failing to leverage its vast internal capacity. This oversight creates two key challenges:

\noindent \textbf{Intra-Layer Redundancy:}
At the weight-level, they ignore that each weight matrix transform input feature space into both \textit{knowledge-sensitive subspaces}, which are crucial for prior tasks, and \textit{redundant subspaces}, which are underutilized yet safe for adaptation. By failing to distinguish between them, existing methods waste valuable capacity that could be repurposed for new learning.

\noindent \textbf{Layer-wise Sensitivity:}
At the layer-level, they overlook that different layers contribute unequally to stability and plasticity, manual selection is risky. Modifying a highly knowledge-sensitive layer can cause catastrophic forgetting, while ignoring an adaptable one wastes an opportunity for new learning.

These challenges lead us to our central research question:
\emph{How can we systematically unlock a model's internal capacity by simultaneously decomposing weights to separate knowledge from redundancy, and by automatically identifying which layers are safest and most effective to adapt?}

To answer this, we introduce \textbf{CKPD-FSCIL}, a unified framework designed to \textbf{open the black box} of pretrained models by reusing internal capacity. It is built upon two innovations to manage knowledge reuse at both the weight and layer levels:

To address the intra-layer redundancy, we first propose a \textbf{Knowledge-Preserving Decomposition (KPD)} module for \textbf{weight-level reuse}. The goal is to partition any linear weight matrix $W$ into two functionally distinct components: a \textbf{frozen, knowledge-sensitive subspace} that protects representations essential to previously learned tasks, and a \textbf{learnable, redundant subspace} that offers safe capacity for adapting to new tasks. 
KPD performs Singular Value Decomposition (SVD) on the product $W\Sigma_\text{in}$, where $\Sigma_\text{in}$ is the input covariance estimated from a small set of past-class samples. This covariance-guided analysis identifies the directions in $W$ most critical for preserving the model's discriminative capabilities.
The magnitude of the singular values guides the subspace partitioning: the components associated with the largest singular values are used to construct the frozen subspace, those associated with the $r$ smallest singular values form the redundant subspace, which is parameterized as a low-rank adapter for training. Only these low-rank adapters are trained. Afterward, they are merged back into the frozen weights, allowing the model to maintain its original structure and incur zero additional inference cost.
We extend this static KPD module into our full \textbf{Continuous KPD (CKPD)} framework by dynamically recalibrating the decomposition before each new session. This is achieved by re-estimating $\Sigma_\text{in}$ from an updated set of past-class samples, ensuring the separation between knowledge and redundancy remains aligned with model's expanding knowledge.

To address the layer-wise sensitivity, we introduce the \textbf{Adaptive Layer Selection (ALS)} mechanism for \textbf{layer-level reuse}. Its purpose is to dynamically identify layers that offer the optimal balance between high adaptation potential and a low risk of catastrophic forgetting.
The core of ALS is the Adapter Sensitivity Ratio (ASR), a principled metric that quantifies this trade-off. ASR evaluates the fraction of a layer's total functional energy contained within its redundant subspace. Adapting such a layer is therefore safe, as changes are confined to a low-impact part of the weight's function. 
At the start of each incremental session, ALS computes the ASR for all layers. , ranks them, and selects the top-$K$ adaptable layers with the lowest ASR scores to apply our KPD module. 
At the start of each incremental session, ALS evaluates the ASR for all linear layers, ranks them accordingly, and selects the top $K$ layers with the lowest ASR scores. These selected layers are then equipped with learnable low-rank adapters, while the rest remain frozen to preserve previously acquired knowledge. 
We extend this to a continuous framework, \textbf{Continuous ALS (CALS)} by repeating the entire selection process at the beginning of each new session using the updated input covariance from the continually growing past-class sample set. This ensures that the allocation of plasticity is not a one-time decision but is dynamically recalibrated, always staying aligned with the model's evolving state.

As shown in Fig.\ref{fig:motivation}~(d), 
these two components work in concert within a select-decompose-train-merge cycle. For each session, CALS first selects the optimal layers, after which CKPD decomposes the weights within them. This allows for targeted training, followed by a zero-overhead merge. By tackling the challenges of weight-level and layer-level knowledge management, our unified framework enables efficient and robust continual learning without architectural modifications or escalating inference costs.
Our contributions are as follows:
\begin{itemize}[leftmargin=*, topsep=2pt, itemsep=0pt]
  \item We propose CKPD-FSCIL, a unified framework that synergistically integrates two mechanisms, Continuous Knowledge-Preserving Decomposition (CKPD) and Continuous Adaptive Layer Selection (CALS), to enable efficient weight-level and layer-level capacity reuse.
  \item We introduce the CKPD mechanism, which partitions any linear layer into frozen and learnable subspaces based on input covariance, enabling targeted adaptation with minimal interference.
  \item We develop the CALS mechanism, powered by our designed adapter sensitivity ratio, to automatically select layers with high adaptation potential and low forgetting risk.
  \item Extensive experiments demonstrate that CKPD-FSCIL achieves state-of-the-art results across multiple FSCIL benchmarks and backbones. Ablations validate the roles of CKPD and CALS, while additional studies confirm its scalability to large-scale ImageNet-1K, generalization to standard CIL, and efficiency in both training and inference.
\end{itemize}
\section{Related Works}
\label{sec:related_works}

\begin{table*}[t!]
\renewcommand\arraystretch{1.2}
\begin{center}
\caption{
\textbf{Comparison of representative FSCIL methods.}
\textbf{P-Free}: No extra trainable parameters at inference.
\textbf{I-Free}: No added computation or latency during inference.
\textbf{S-Free}: No additional supervision required.
\textbf{Layer Sel.}: Whether layer selection is adaptive.
}
\vspace{-2mm}
\resizebox{\linewidth}{!}{
\begin{tabular}{lp{9cm}cccc}
\toprule
\textbf{Method} & \textbf{Description} & \makecell{\textbf{P-Free} \\ (Params)} & \makecell{\textbf{I-Free} \\ (Infer)} & \makecell{\textbf{S-Free} \\ (Sup)} & \makecell{\textbf{Layer} \\ \textbf{Sel.}} \\
\midrule
CPE-CLIP~\cite{d2023multimodal} & Adds learnable multimodal prompts to both the vision and language branches of CLIP, alongside a regularization loss to stabilize training. & \xmark & \xmark & \xmark & Manual \\
PriViLege~\cite{park2024pre} & Introduces base and vision-language prompts to enable cross-session knowledge transfer, and incorporates entropy-based divergence loss and knowledge distillation from a pretrained language model. & \xmark & \xmark & \xmark & Manual \\
PL-FSCIL~\cite{tian2024pl} & Applies domain- and task-specific visual prompts on a pretrained ViT, with orthogonality constraints enforced through a prompt regularization loss. & \xmark & \xmark & \xmark & Manual \\
ASP-FSCIL~\cite{liu2024few} & Employs attention-aware, self-adaptive prompts to model shared and specific knowledge, guided by task-invariant/specific prompt separation and an additional information bottleneck loss. & \xmark & \xmark & \xmark & Manual \\
FSPT-FSCIL~\cite{ran2024brain} & Uses fast- and slow-updating prompts trained via meta-learning to balance plasticity and stability. & \xmark & \xmark & \cmark & Manual \\
KANet~\cite{wang2024knowledge} & Inserts knowledge adapter modules into the network to integrate data-specific cues into general representations. & \xmark & \xmark & \cmark & Manual \\
\midrule
\rowcolor{mygray}CKPD-FSCIL & Reuses the model’s internal capacity by decomposing and selectively adapting pretrained weights, enabling efficient and stable continual learning without architectural changes or inference overhead. & \cmark & \cmark & \cmark & Adaptive \\
\bottomrule
\end{tabular}
}
\label{tab:compare_methods}
\end{center}
\vspace{-6mm}
\end{table*}

\subsection{Few-shot Class-Incremental Learning}
FSCIL~\cite{tao2020few} unifies the challenges of Class-Incremental Learning (CIL)~\cite{rebuffi2017icarl, li2017learning, zhang2023few, tian2024survey} and Few-Shot Learning (FSL)~\cite{vinyals2016matching, ravi2017optimization}, requiring models to learn new classes from limited samples in each session without accessing past data. This setting is prone to two key issues: (1) \textit{Catastrophic forgetting}~\cite{mccloskey1989catastrophic, goodfellow2013empirical, yang2024corda}, where learning new classes overwrites previously acquired knowledge; (2) \textit{Overfitting to scarce samples}~\cite{snell2017prototypical, sung2018learning}, which limits generalization to novel categories. Addressing the \textit{stability-plasticity dilemma}~\cite{mermillod2013stability} necessitates balancing \textit{stability} (preserving prior knowledge) and \textit{plasticity} (learning new concepts).

To this end, various strategies have been explored. \textit{Replay-based methods}~\cite{liu2022few,agarwal2022semantics,peng2022few} store a small exemplar buffer or synthesize pseudo-data to mitigate forgetting.
\textit{Optimization-based methods} improve generalization via meta-learning~\cite{yoon2020xtarnet,chi2022metafscil,zhou2022few}, pseudo-feature generation~\cite{cheraghian2021synthesized,zhou2022forward}, advanced regularization~\cite{tao2020topology,joseph2022energy,lu2022geometer,chen2021incremental,akyurek2021subspace,song2023learning,ahmed2024orco,zou2024compositional,goswami2024calibrating,zhou2024delve,oh2024closer}, improved classifiers~\cite{hersche2022constrained,yang2023neural,yang2023neural_arxiv,zhu2021self,wang2024few}, and knowledge distillation~\cite{tao2020few,dong2021few,cheraghian2021semantic,zhao2023few}. 
While effective, these approaches are largely complementary to the fundamental architectural question of how a model's parameters should be adapted during continual learning. We focus on innovating the core parameter adaptation paradigm to achieve an optimal stability-plasticity balance that requires no architectural changes or complex loss terms.

\vspace{-2mm}\subsection{Parameter Adaptation Strategies}
\label{sec:relatedwork_parameter_adaptation}
The approach to parameter adaptation is central to solving the FSCIL problem. Existing methods can be broadly categorized into three dominant paradigms:

\vspace{1mm} \noindent \textbf{Full Fine-tuning vs. Full Freezing.}
Some methods fine-tune the entire network, offering high plasticity but suffering from severe forgetting and overfitting~\cite{rebuffi2017icarl, hou2019learning, kirkpatrick2017overcoming, shin2017continual}. Conversely, the most common strategy is to completely freeze the feature extractor to ensure perfect stability, but this severely limits the model's ability to learn new discriminative features~\cite{zhang2021few, yang2023neural, li2024mamba}.

\vspace{1mm} \noindent \textbf{Prompt/Adapter Tuning.}
To balance stability and plasticity, recent methods employ parameter-efficient tuning by freezing the backbone and injecting lightweight, learnable modules. These strategies fall into two categories:
\textit{Prompt-based tuning} prepends learnable vectors (prompts) to inputs or intermediate representations~\cite{lester2021power, li2021prefix, jia2022visual, wang2022dualprompt}, as seen in FSCIL methods like CPE-CLIP~\cite{d2023multimodal} and PriViLege~\cite{park2024pre}.
\textit{Adapter-based tuning} inserts trainable modules between frozen layers~\cite{houlsby2019parameter, he2022towards, lei2023conditional,hu2022lora}. FSCIL variants such as KANet~\cite{wang2024knowledge} and and CA-CLIP~\cite{goswami2024calibrating} use adapters for task-specific adaptation.
While effective, these methods face several critical:
\textbf{(1) Parameter/Inference overhead.} Most methods introduce new modules per session, leading to growing model size and inference costs.
\textbf{(2) Extra supervision.} Some approaches require complex training objectives, such as knowledge distillation or regularization constraints, increasing the need for extra supervision.
\textbf{(3) Underutilized pretrained weights.} They treat the pretrained weights as black box, failing to leverage its internal capacity. This limitation manifests at two levels: \textbf{At the weight-level}, they ignore the redundant subspaces within existing weights, forcing the model to rely on external modules for new knowledge and thus wasting valuable internal plasticity. \textbf{At the layer-level}, adapters are inserted heuristically, lacking an mechanism to identify which layers are safest or most effective to modify.
To clarify these limitations, Tab.~\ref{tab:compare_methods} summarizes representative FSCIL methods.

We address these challenges with three key innovations:
\textbf{(1) Zero-overhead weight reuse.} We repurpose redundant subspaces within existing weights to form learnable adapters. These are merged back after training, ensuring zero additional parameters or inference cost while preserving the original architecture.
\textbf{(2) No extra supervision.} Our covariance-guided decomposition intrinsically preserves knowledge, eliminating the need for auxiliary losses or external supervision.
\textbf{(3) Adaptive layer selection:} We automatically select layers with high plasticity and low forgetting risk, eliminating the need for manual heuristics.

\vspace{1mm} \noindent \textbf{Decomposition-Based Adaptation.}
Low-Rank Adaptation (LoRA)~\cite{hu2022lora} pioneer a mergeable low-rank update strategy that eliminates inference overhead by integrating learned adapters into pretrained weights. While effective in reducing trainable parameters, LoRA’s random initialization is data-agnostic, leaving the pretrained weights as a black box and failing to exploit their internal representational structure.
PiSSA~\cite{meng2024pissa} addresses this by applying SVD directly to the raw weight matrix $W$ into principal and residual components to initialize low-rank adapters. However, PiSSA remains data-agnostic, as it ignores how $W$ interacts with input data. ASVD~\cite{yuan2023asvd} advances this by performing SVD on the activation-weighted product $W\gF$, where $\gF$ is the activation matrix, allowing the decomposition to reflect the model’s functional behavior. This activation-guided design inspired subsequent works such as CorDA~\cite{yang2024corda}, which replaces raw activations with a covariance matrix derived from context-representative samples, orienting the decomposition toward a specific downstream task. In parallel, DoRA~\cite{liu2024dora} adopts a different philosophy, decomposing $W$ into magnitude and direction components and applying LoRA-like updates only to the direction, effectively changing the adaptation target rather than the initialization process.

Despite these advances, prior methods remain confined to static, single-task fine-tuning and fall short in two crucial aspects for FSCIL:
\textbf{(1) Continual knowledge integration}. Their decompositions are performed only once and remain fixed, ignoring the evolving nature of incremental tasks. This prevents the model from effectively incorporating new knowledge while preserving old.
\textbf{(2) Lack of principled layer selection}. Decomposition is applied uniformly or based on heuristics, without considering the varying sensitivity of different layers to forgetting. This risks either unnecessary rigidity or harmful interference.

Our CKPD-FSCIL is the first to extend weight decomposition to vision-based FSCIL with both \textbf{continuous recalibration} of the decomposition using up-to-date covariance statistics each session, and \textbf{adaptive layer selection} via a quantitative sensitivity measure. These components transform one-time, static decomposition into a dynamic, layer-aware framework that preserves prior knowledge, adapts efficiently to new tasks, and incurs zero inference overhead.
\section{Method}
\label{sec:method}
\begin{figure*}[t]
\centering
\includegraphics[width=1\linewidth]{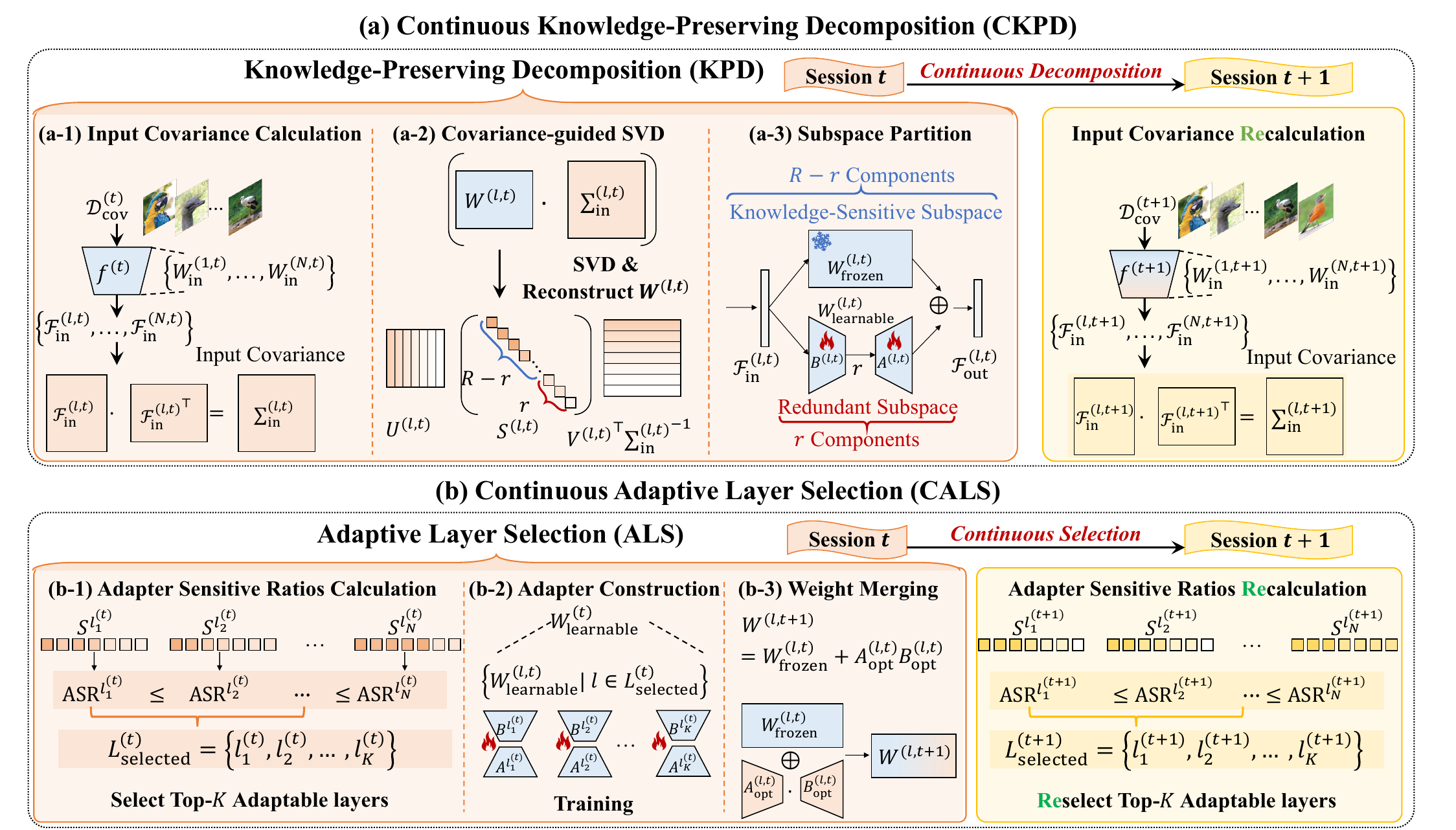}
\caption{\textbf{Overview of the CKPD-FSCIL framework}: Our method integrates two core mechanisms for continual adaptation.
\textbf{(a) Continuous Knowledge-Preserving Decomposition} performs weight-level adaptation on a selected layer $W^{(l,t)}$ at layer $l$ during session $t$:
(1) Input Covariance Calculation: Compute the input covariance $\Sigma_\text{in}^{(l,t)}$ from a small covariance sample set $D_\text{cov}$.
(2) Covariance-Guided Decomposition: Apply SVD to $W^{(l,t)}\Sigma_\text{in}^{(l,t)}$ to analyze the weight's functional importance.
(3) Subspace Partition: Split $W^{(l,t)}$ into a frozen knowledge-sensitive subspace ($W_\text{frozen}$) reconstructed from the components associated with the top $R-r$ singular values, and a learnable, low-rank redundant subspace ($W^{(l,t)}_\text{learnable} = B^{(l,t)} A^{(l,t)}$) reconstructed from the components associated with the smallest $r$ singular values.
The process is made continuous by recalculating $\Sigma_\text{in}^{(l,t+1)}$ at the start of the next session $t+1$ to reflect the expanded knowledge base.
\textbf{(b) Continuous Adaptive Layer Selection} shows the layer-level selection process:
(4) ASR Calculation: The adapter sensitivity ratios are computed for all $N$ linear layers to quantify their adaptation potential.
(5) Adapter Construction \& Training: Adapters are constructed only for the $K$ layers with the lowest ASR scores ($L_\text{learnable}^{(t)}$), which are the only components trained in the backbone.
(6) Weight Merging: After training, the optimized adapter ($B^{(l,t)}_\text{opt}, A^{(l,t)}_\text{opt}$) is merged back into $W^{(l,t)}_\text{frozen}$.
This selection process is also continuous, with ASRs being recomputed and the set of learnable layers $L_\text{selected}^{(t+1)}$ reselected at the start of the next session $t+1$.
}
\label{fig:framework}
\vspace{-4mm}
\end{figure*}

Our CKPD-FSCIL framework addresses the core challenges of FSCIL by introducing a parameter adaptation paradigm that repurposes a model’s internal capacity without adding external modules. It resolves the stability–plasticity dilemma through two complementary components: the continuous knowledge-preserving decomposition, which determines what to adapt at the weight level, and the continuous adaptive layer selection, which decides which layers to adapt across the network. We begin with the problem formulation (Sec.~\ref{sec:method_problem_formulation}), followed by a detailed description of each component (Secs.~\ref{sec:method_ckpd}, \ref{sec:method_cals}) and their integration into a unified, cyclical adaptation process (Sec.~\ref{sec:method_unified_framework}).

\vspace{-2mm}\subsection{Problem Formulation}
\label{sec:method_problem_formulation}
FSCIL trains a model incrementally over multiple sessions, denoted as $\{\gD^{(0)}, \gD^{(1)}, \dots, \gD^{(T)}\}$. In each session $t$, the model receives a training set $\gD^{(t)} = \{(\vx_i, y_i)\}_{i=1}^{|\gD^{(t)}|}$, where $\vx_i$ is an input sample and $y_i$ is its corresponding label. The base session $\gD^{(0)}$ provides a comprehensive label set $\gC^{(0)}$ with substantial data for each class, serving as the foundation for the model's initial learning. In subsequent sessions $\gD^{(t)}$, $t > 0$, the model learns new classes with only a few labeled examples per class, typically following a $p$-way $q$-shot setup—meaning $p$ new classes with $q$ samples for each class. There is no overlap between the classes of different sessions, \emph{i.e.}, $\gC^{(t)} \cap \gC^{(t')} = \emptyset$ for all $t' \neq t$. Other than data replay with limited samples, the training data from previous sessions are inaccessible in future sessions. During evaluation in session $t$, the model is tested on data from all classes encountered, \emph{i.e}, $\bigcup_{i=0}^{t} \gC^{(i)}$. The goal is to achieve high accuracy across all learned classes, balancing the acquisition of new knowledge with the retention of existing knowledge.

\vspace{-2mm}\subsection{Continuous Knowledge-Preserving Decomposition}
\label{sec:method_ckpd}
To enable weight-level knowledge reuse, we introduce the \textbf{Continuous Knowledge-Preserving Decomposition (CKPD)} mechanism (Fig.~\ref{fig:framework}~(a)) to open the black box of a single linear layer, which consists of two parts:
\textbf{(1) Knowledge-preserving decomposition (KPD):} Covariance-guided decomposition splits each layer’s weight matrix into a frozen, knowledge-sensitive subspace for preserving prior knowledge and a learnable, redundant subspace for learning new information.
\textbf{(2) Continual adaptation:} At the start of each incremental session, the decomposition is recalibrated to align with the model’s evolving knowledge base.

In the following, we detail the theoretical motivation for our KPD (Sec.~\ref{sec:method_ckpd_motivation}), the step-by-step implementation of the single-session KPD module (Sec.~\ref{sec:method_ckpd_kpd}), and finally, how it is extended to the CKPD setting (Sec.~\ref{sec:method_ckpd_continuous_decomposition}).

\subsubsection{Theoretical Motivation}
\label{sec:method_ckpd_motivation}
Our decomposition design stems from the insight that \textit{not all directions in a weight matrix contribute equally to classification}. Each weight matrix $W$ comprises a \textbf{knowledge-sensitive subspace} that generates high-variance, task-discriminative outputs, and a \textbf{redundant subspace} with limited influence on prior tasks that can be repurposed for new learning. We aim to preserve the former to mitigate forgetting and adapt the latter for continual learning.
A straightforward approach of applying Singular Value Decomposition (SVD) directly on $W$ is \textbf{data-agnostic}, revealing only its intrinsic structure without considering how it transforms inputs into discriminative outputs.

To achieve a \textbf{data-guided} decomposition, we define “knowledge” as the geometric structure of the output feature space shaped by training. From classical discriminant analysis to recent deep learning theory~\cite{fisher1936use, mclachlan2005discriminant, he2023law,yuan2023asvd,yang2024corda}, it is well established that a well-trained model maps inputs to an output space with \textit{high inter-class variance} and \textit{low intra-class variance}. The principal components of the output covariance matrix $\Sigma_\text{out}$ capture the most discriminative directions, making their preservation essential to prevent forgetting.

To understand the formation of this discriminative structure, consider a linear transformation $y = Wx$. The output covariance is:
\vspace{-2mm}
\begin{equation}
\Sigma_\text{out} = \mathbb{E}[yy^\top]
         = \mathbb{E}[(Wx)(Wx)^\top]
         = W\mathbb{E}[xx^\top]W^\top 
         = W\Sigma_\text{in} W^\top,
    \label{eq:cov_transform}
\end{equation}
where $\Sigma_\text{in} = \mathbb{E}[xx^\top]$ is the input data's covariance.
Thus, the output covariance $\Sigma_\text{out}$, which governs the model’s discriminative capacity, depends jointly on $W$ and the input covariance $\Sigma_\text{in}$.
Therefore, to identify which directions in $W$ are most knowledge-sensitive, we analyze its interaction with $\Sigma_\text{in}$ by performing SVD on the product $W\Sigma_\text{in}$, which reflects how $W$ amplifies different input directions. Performing SVD on $W\Sigma_\text{in}$ and using the resulting singular values allows us to retain the top variance-preserving components while repurposing the remaining subspace for learning new tasks.

\subsubsection{Knowledge-Preserving Decomposition}
\label{sec:method_ckpd_kpd}
KPD partitions a linear layer’s weight matrix $W$ into a frozen, knowledge-sensitive subspace ($W_\text{frozen}$) and a learnable, redundant subspace ($W_\text{learnable}$) through following steps:

\vspace{1mm} \noindent \textbf{Input Covariance Calculation.}
To guide the decomposition, we compute the input feature covariance matrix using a past-class sample set $\gD_{\text{cov}}^{(t)}$, which contains \textit{one randomly selected sample per previously seen class}:
\vspace{-4mm}
\begin{equation}
\gD_{\text{cov}}^{(t)} = \{(\vx_{c}, y_{c}) \mid y_{c} \in \bigcup_{i=0}^{t-1} \gC^{(i)},\ \vx_{c} \sim \operatorname{Rand}(\gD_{y_c}, 1) \},
\end{equation}
where $\gD_c$ is the set of all available samples for class $y_c$, and $\operatorname{Rand}(\gD_c, 1)$ returns a single, randomly chosen sample. Although using a single sample per class appear minimal, we empirically demonstrate in Sec.~\ref{sec:exps_generalization_random_seeds} that it is sufficient to reliably capture the covariance structure needed for stable performance.

At the start of each session $t > 0$, we forward $\gD_{\text{cov}}^{(t)}$ through the backbone $f$ to extract token-level activations before each linear layer $W^{(l,t)} \in \mathbb{R}^{d_\text{out} \times d_\text{in}}$, where $l$ denotes the layer index, $d_\text{in}$ is the input dimension, and where $d_\text{out}$ is the output dimension. As shown in Fig.~\ref{fig:framework}~(a-1), the resulting input features for $W^{(l,t)}$ is $\gF_\text{in}^{(l,t)} \in \mathbb{R}^{d_{\text{in}} \times (N_\text{cov} \times N_{\text{patch}}})$, where $N_{\text{cov}} = |\gD_{\text{cov}}^{(t)}|$ is the number samples in $\gD_{\text{cov}}^{(t)}$; $N_\text{patch}$ is the number of tokens per input. 

Assuming features are centered by Layer Normalization operation, we compute the input covariance matrix for $W^{(l,t)}$ as:
\begin{equation}
\vspace{-2mm}
\Sigma_\text{in}^{(l,t)} = \frac{1}{N_{\text{cov}} \cdot N_{\text{patch}}} \gF_\text{in}^{(l,t)} {\gF_\text{in}^{(l,t)}}^\top \in \mathbb{R}^{d_{\text{in}} \times d_{\text{in}}}.
\label{eq:input_covariance}
\end{equation}
This computation is applied independently to each linear layer. $\Sigma_\text{in}$ encodes dominant directions of prior knowledge and serves as the statistical prior for the subsequent decomposition step.

\vspace{1mm} \noindent \textbf{Covariance-Guided SVD.}
To reveal how the weight matrix $W^{(l,t)}$ interacts with the input distribution, we apply SVD to the covariance-transformed weight, $W^{(l,t)}\Sigma_\text{in}^{(l,t)}$ (Fig.~\ref{fig:framework}~(a-2)):
\begin{equation}
    \text{SVD}~(W^{(l,t)}\Sigma_\text{in}^{(l,t)}) = U^{(l,t)} S^{(l,t)} {V^{(l,t)}}^\top = \sum_{i=1}^{R} s_i \mathbf{u}_i \mathbf{v}_i^\top,
    \label{equ:kpd}
\end{equation}
where $U^{(l,t)} \in \mathbb{R}^{d_{\text{out}} \times d_{\text{out}}}$ and $V^{(l,t)} \in \mathbb{R}^{d_{\text{in}} \times d_{\text{in}}}$ are orthogonal matrices, with the $i$-th left singular vector $\mathbf{u}_i \in \mathbb{R}^{d_{\text{out}}}$ and the $i$-th right singular vector $\mathbf{v}_i \in \mathbb{R}^{d_{\text{in}}}$ corresponding to the $i$-th columns of $U^{(l,t)}$ and $V^{(l,t)}$, respectively. The singular value matrix $S^{(l,t)} \in \mathbb{R}^{d_\text{out} \times d_\text{in}}$ is rectangular diagonal, where $R = \min{d_\text{out}, d_\text{in}}$ denotes the rank of $W^{(l,t)}$. Its top-left $R \times R$ block contains the singular values $s_1 \geq \cdots \geq s_R > 0$ in descending order, with all other entries set to zero.

To map the decomposition from the covariance-adjusted space back to the original weight space, and crucially, to prevent any model drift at the start of each new adaptation session, we use $W= (W\Sigma_\text{in}) (\Sigma_\text{in})^{-1} = U S {V}^\top (\Sigma_\text{in})^{-1} $ to reconstructing $W$ as:
\begin{equation}
\begin{aligned}
 W^{(l,t)} =  U^{(l,t)} S^{(l,t)} {(V^{(l,t)}}^\top {\Sigma_\text{in}^{(l,t)}}^{-1}) =  \sum_{i=1}^{R} s_i \mathbf{u}_i \hat{\mathbf{v}}_i^\top, 
\end{aligned}
\label{eq:decomposed-reconstruct}
\end{equation}
where $(\Sigma_\text{in}^{(l,t)})^{-1}$ denotes the inverse of $\Sigma_\text{in}^{(l,t)}$, and $\hat{\mathbf{v}}_i^\top = \mathbf{v}_i^\top (\Sigma_\text{in}^{(l,t)})^{-1}$ denotes the covariance-adjusted input direction in the original feature space. 
To ensure invertibility of $\Sigma_\text{in}$, we dynamically regularize $\Sigma_\text{in}$ when it is not invertible. Specifically, we add a scaled identity matrix to the diagonal of $\Sigma_\text{in}$ via $\tilde{\Sigma}_x = \Sigma_\text{in} + \lambda \cdot \text{mean}~(\text{diag}~(\Sigma_\text{in})) \cdot I$, where $\lambda$ is a small positive coefficient. We then compute the reconstruction error $| \tilde{\Sigma}_x \tilde{\Sigma}_x^{-1} - I |_2$. If the error exceeds a threshold, we double $\lambda$ and repeat this process until the deviation falls below the threshold.

\vspace{1mm} \noindent \textbf{Subspace Partition and Adapter Construction.}
As shown in Eq.~\ref{eq:decomposed-reconstruct}, $W$ can be expressed as a weighted sum of directional components, each mapping an adjusted input direction $\hat{\mathbf{v}}_i$ to an output direction $\mathbf{u}_i$, scaled by its singular value $s_i$.
The magnitude of $s_i$ quantifies how strongly $\hat{\mathbf{v}}_i$ influences the output representation along $\mathbf{u}_i$, thus reflecting that component’s importance: large $s_i$ values capture directions crucial for maintaining inter-class variance and preserving prior knowledge, whereas small $s_i$ values correspond to less critical directions that can be reassigned for new tasks.
Guided by this principle, we partition the weight space into two complementary subspaces:
\begin{itemize}[leftmargin=*, topsep=2pt, itemsep=0pt]
    \item \textit{Knowledge-Sensitive Subspace.} Defined by the components corresponding to the top $R - r$ singular values $S_{\text{frozen}} = \operatorname{diag}~(s_1, \dots, s_{R - r})$, which are critical for preserving knowledge acquired from previous tasks.
    \item \textit{Redundant Subspace.} Defined by the components corresponding to the smallest $r$ singular values $S_{\text{learnable}} = \operatorname{diag}~(s_{R - r + 1}, \dots, s_R)$, which contribute minimally to prior knowledge and can be safely repurposed for learning new tasks.
\end{itemize}

We represent the \textit{redundant subspace} $W_\text{learnable}^{(l,t)} \in \mathbb{R}^{d_{\text{out}} \times d_{\text{in}}}$ using bottom-$r$ singular components, as shown in Fig.~\ref{fig:framework}~(a-3):
\begin{equation}
\begin{aligned}
W_\text{learnable}^{(l,t)} &= U^{(l,t)} S_\text{learnable}^{(l,t)} ({V^{(l,t)}}^\top {\Sigma_\text{in}^{(l,t)}}^{-1}) 
= \sum_{i=R-r+1}^{R} s_i \mathbf{u}_i \mathbf{\hat{v}}_i^\top.
\end{aligned}
\end{equation}

To enable efficient adaptation, we parameterize $W_\text{learnable}$ using a low-rank factorization:
\begin{equation}
\label{equ:W_learable}
    W_\text{learnable}^{(l,t)} = B^{(l,t)}A^{(l,t)},
\end{equation}
where $B^{(l,t)} \in \mathbb{R}^{d_{\text{out}} \times r}$ and $A^{(l,t)} \in \mathbb{R}^{r \times d_{\text{in}}}$ are only trainable parameters during adaptation. This low-rank form offers two benefits:
\begin{itemize}[leftmargin=*, topsep=2pt, itemsep=0pt]
    \item \textit{Parameter Efficiency.} Directly optimizing $W_\text{learnable}$ as a dense matrix ($d_{\text{out}} \times d_{\text{in}}$) is inefficient, as its inherent rank $r$ is much smaller than $\min(d_{\text{out}}, d_{\text{in}})$. The low-rank form reduces trainable parameters from $O(d_{\text{out}} \times d_{\text{in}})$ to $O(r \cdot (d_{\text{out}}+ d_{\text{in}}))$, leading to faster training and lower memory overhead.
    \item \textit{Regularization.} Constraining updates within a compact low-dimensional subspace implicitly regularizes the model. This helps mitigate overfitting, especially in few-shot settings where labeled data is limited.
\end{itemize}

The low-rank adapter is initialized by symmetrically factorizing the singular value matrix of the redundant subspace. This distributes its square roots across $B$ and $A$, which ensures a balanced gradient flow and promotes stable training:
\begin{equation}
\label{equ:BA}
\begin{aligned}
    & B^{(l,t)} = U^{(l,t)}_{[:,R-r:]} \sqrt{S^{(l,t)}_\text{learnable}}\in \mathbb{R}^{d_{\text{out}} \times r}, \\
    & A^{(l,t)} = \sqrt{S^{(l,t)}_\text{learnable}} ({V^{(l,t)}}^\top {\Sigma_\text{in}^{(l,t)}}^{-1})_{[R-r:,:]} \in \mathbb{R}^{r \times d_{\text{in}}},
\end{aligned}
\end{equation}
where $U^{(l,t)}_{[:, R - r:]}$ and $({V^{(l,t)}}^\top {\Sigma_\text{in}^{(l,t)}}^{-1})_{[R - r:, :]}$ denote the last $r$ columns and rows of the respective matrices.

We define the frozen \textit{knowledge-sensitive subspace} $W_{\text{frozen}} \in \mathbb{R}^{d_{\text{out}} \times d_{\text{in}}}$ to preserve previously acquired knowledge. A crucial requirement for the decomposition is that splitting $W$ into $W_{\text{frozen}}$ and $W_{\text{learnable}}$ must not alter the model's output before any adaptation. That is, for the input $\gF_\text{in}^{(l,t)}$ of $W^{(l,t)}$, the original prediction result $\gF_\text{out}^{(l,t)}$ must remain unchanged:
\begin{equation}
\gF_\text{out}^{(l,t)} = W^{(l,t)}\gF_\text{in}^{(l,t)} = W_{\text{frozen}}^{(l,t)} \gF_\text{in}^{(l,t)} + W_{\text{learnable}}^{(l,t)}\gF_\text{in}^{(l,t)}.
\end{equation}

Theoretically, $W_{\text{frozen}}$ could be reconstructed from the top $R-r$ singular components via $W_{\text{frozen, ideal}} = U S_\text{frozen} (V^\top{\Sigma_\text{in}^{(l,t)}}^{-1}) =\sum_{i=1}^{R-r} s_i \mathbf{u}_i \mathbf{\hat{v}}_i^\top$. However, this process involves SVD and matrix inversion, which may introduce minor floating-point errors. Though small, such discrepancies can accumulate in high-dimensional weights, causing $W_{\text{frozen}} + W_{\text{learnable}} \neq W$. As a result, predictions may slightly shifts even before training, leading to minimal performance drop and knowledge forgetting.

To guarantee that the original weight matrix can be exactly recovered through $W^{(l,t)}= W_{\text{frozen}} + W_{\text{learnable}}$, we define $W_{\text{frozen}}$ directly as the residual:
\begin{equation}
W_{\text{frozen}}^{(l,t)} = W^{(l,t)}- W_{\text{learnable}}^{(l,t)} = W^{(l,t)}- B^{(l,t)}A^{(l,t)}.
\end{equation}
This ensures the model's inference outputs remain strictly unchanged before and after decomposition, providing prior knowledge stability without any initial prediction shift.
During training, $W_{\text{frozen}}$ remains fixed, confining updates to the redundant subspace and minimizing interference with previously acquired knowledge.

\vspace{1mm} \noindent \textbf{Weight Merging.}
After the adaptation phase, the optimized adapter parameters $B^{(l,t)}_\text{opt}$ and $A^{(l,t)}_\text{opt}$ are merged back into the frozen component, restores the original architecture and incurs no inference overhead, as shown in Fig.~\ref{fig:framework}~(b-3).
The final updated weight matrix for the new session $t+1$ is:
\begin{equation}
    W^{(l,t+1)} = W^{(l,t)}_\text{frozen} + B^{(l,t)}_\text{opt} A^{(l,t)}_\text{opt},
\end{equation}

\subsubsection{Continuous Decomposition: From KPD to CKPD}
\label{sec:method_ckpd_continuous_decomposition}
While KPD prevents catastrophic forgetting, the preserved subspace remains unchanged after the base session, gradually becoming misaligned with the evolving feature distribution and unable to assimilate newly learned knowledge. 
To ensure our framework remains effective throughout the entire continual learning process, we extend the one-shot KPD into a continuous process, which we term \textbf{Continuous KPD (CKPD)}.
CKPD \emph{re-evaluates and updates} the preserved subspace at the start of each incremental session, guided by the latest feature covariance. This dynamic refresh ensures that the preserved component always captures stable yet up-to-date knowledge, while the redundant subspace is safely allocated for new class integration.

At the beginning of each new incremental session $t+1$, we first update the covariance sample set by adding one randomly selected exemplar for each newly introduced class $\gC^{(t)}$:
\begin{equation}
\gD_\text{cov}^{(t+1)} = \gD_\text{cov}^{(t)} \cup \{ (\vx_{c}, y_{c}) \mid y_{c} \in \gC^{(t)},\ \vx_{c} \sim \operatorname{Rand}(\gD_{y_c})\}.
\end{equation}

Using the updated sample set, we recalculate the session-specific input covariance matrix, $\Sigma_\text{in}^{(l,t+1)}$, which is then used for a new KPD on each selected weight to obtain an updated frozen subspace  $W_{\text{frozen}}^{(l,t+1)}$, and a newly initialized learnable adapter, $W_{\text{learnable}}^{(l,t+1)} = B^{(l,t+1)} A^{(l,t+1)}$.
By repeating this recalibration process before each session, CKPD ensures that the partitioning between knowledge-sensitive and redundant subspaces is always aligned with the model's complete and most up-to-date knowledge base.

\vspace{-2mm}\subsection{Continuous Adaptive Layer Selection}
\label{sec:method_cals}
While CKPD enables safe adaptation of individual layers, applying it uniformly is inefficient, as layers differ in sensitivity to prior knowledge. Modifying a highly sensitive layer risks forgetting, while ignoring a highly adaptable one wastes learning potential.  
We introduce the \textbf{Continuous Adaptive Layer Selection (CALS)} mechanism  to automatically choose layers with the best trade-off between adaptation potential and forgetting risk (Fig.~\ref{fig:framework}~(b)).

\subsubsection{Adapter Sensitivity Ratio (ASR)}
\label{sec:method_cals_asr}
We achieve the CALS mechanism with a new proposed metric called the Adapter Sensitivity Ratio (ASR), which quantifies how much of the layer’s representational capacity is allocated to the learnable subspace. A small ratio indicates that the adaptable part is truly insignificant, making it a safe candidate for modification.

For a given layer $l$, let $S^{(l)} = (s_1^{(l)}, s_2^{(l)}, \dots, s_R^{(l)})$ be the singular values from KPD (Eq.~\ref{equ:kpd}), sorted in descending order, as shown in Fig.~\ref{fig:framework}~(b-1). Given an adapter rank $r$, we define the ASR as:
\begin{equation}
\text{ASR}^{(l)} = \frac{\sum_{i=R - r + 1}^{R} s_i^{(l)}}{\sum_{i=1}^{R} s_i^{(l)}},
\label{eq:asr}
\end{equation}
where the numerator represents the energy of the redundant subspace (the sum of the smallest $r$ singular values), while the denominator represents the total energy of all components.

\textit{A low ASR} indicates that the redundant subspace contributes minimally to the layer's overall function. Adapting this layer is considered safe, as changes are confined to a low-impact part of the weight's function.
\textit{A high ASR} suggests that even the least important directions represent a substantial portion of the layer's energy. This implies a blurry boundary between knowledge and redundancy, making the layer sensitive and risky to adapt.
As a normalized ratio, ASR allows to compare the sensitivity across different layers with different energy scales.

\subsubsection{Layer Selection Procedure}
\label{sec:method_cals_procedure}
At the beginning of each incremental session $t$, we dynamically select the top-$K$ most adaptable layers. This procedure is continuous, as it is based on the session-specific covariance $\Sigma_\text{in}^{(l,t)}$. Our selection procedure is as follows:

\begin{itemize}[leftmargin=*, topsep=2pt, itemsep=0pt]
    \item \textbf{Compute ASR Scores:} For each linear layer $l \in \{1, \dots, N\}$, we compute its ASR score, $\text{ASR}^{(l,t)}$, using the singular values $S^{(l,t)}$ from our KPD process ( Eq.~\ref{eq:asr}).
    \item \textbf{Rank Layers:} We rank all layers in ascending order based on their ASR scores. Let $l^{(t)} = (l_1^{(t)}, \dots, l_N^{(t)})$ be the sorted list of layer indices:
    \begin{equation}
        \text{ASR}^{l_1^{(t)}} \leq \text{ASR}^{l_2^{(t)}} \leq \dots \leq \text{ASR}^{l_N^{(t)}}.
    \end{equation}
    \item \textbf{Select Top-K Adaptable Layers:} We select the $K$ layers with the lowest ASR scores as the set of adaptable layers for session $t$:
    \begin{equation}
    L_\text{selected}^{(t)} = \left\{ l_1^{(t)}, l_2^{(t)}, \dots, l_K^{(t)} \right\}.
    \label{eq:selected_layers}
    \end{equation}
    \item \textbf{Adapter Construction:} The KPD process is then applied only into the selected layers $\gL_{\text{selected}}^{(t)}$, updating their redundant subspaces $W_\text{learnable}$ while keeping the knowledge-sensitive components $W_\text{frozen}$ and the remaining $N-K$ layers fixed. The session-specific learnable weights Fig.~\ref{fig:framework}~(b-2), are:
    \begin{equation}
    W_\text{learnable}^{(t)} = \left\{ W_\text{learnable}^{(l,t)} \mid l \in L_\text{selected}^{(t)} \right\}.
    \label{eq:selected_weights}
    \end{equation}
where $W_\text{learnable}^{(l,t)}$ denotes the learnable adapter of layer $l$ instantiated for session $t$. This dynamic allocation of plasticity procedure is repeated in each session using the updated covariances, ensuring adaptation remains both safe and effective.
\end{itemize}

\vspace{-2mm}\subsection{The Unified CKPD-FSCIL Framework}
\label{sec:method_unified_framework}
The CKPD and CALS mechanisms are integrated into our CKPD-FSCIL framework, which is a unified, cyclical solution for the stability-plasticity dilemma. As depicted in Fig~\ref{fig:framework}, for each incremental session $t$, CKPD-FSCIL executes a select-decompose-train-merge pipeline:
\begin{enumerate}[leftmargin=*, topsep=2pt, itemsep=0pt]
    \item \textbf{Select}: At the beginning of the session, CALS first analyzes all linear layers. It computes the ASR for each and selects the top-$K$ layers with the lowest scores, identifying them as the safest and most effective candidates for adaptation (Sec.~\ref{sec:method_cals}).
    \item \textbf{Decompose}: For each of the layers selected by CALS, CKPD is then applied. It partitions the layer's weight matrix into a frozen, knowledge-sensitive component and a learnable, low-rank adapter that represents the redundant subspace (Sec.~\ref{sec:method_ckpd}). All other layers remain entirely frozen.
    \item \textbf{Train}: During training, updates are exclusively confined to the newly created low-rank adapters in the selected layers. This targeted adaptation minimizes interference with the model's existing knowledge base, which is protected by both the frozen components and the untouched layers.
    \item \textbf{Merge}: After training, the optimized adapters are merged back into their corresponding frozen components. This step restores the original model architecture, ensuring that our method incurs zero additional parameters or inference overhead in final model.
\end{enumerate}
This entire process is repeated for each new incremental session, with the covariance information and ASR scores being dynamically recalculated. This ensures that the allocation of plasticity is continuously recalibrated, always staying aligned with the model's evolving state and providing a robust, efficient, and truly unified framework for few-shot continual learning.

\section{Experiments}
\label{sec:exps}
\subsection{Implementation Details}
\label{sec:exps_details}
\noindent \textbf{Datasets.} We evaluate CKPD-FSCIL on three standard FSCIL benchmarks: miniImageNet~\cite{russakovsky2015imagenet}, CUB-200~\cite{wah2011caltech}, and CIFAR-100~\cite{krizhevsky2009learning}. For all datasets, we follow the widely adopted protocols from prior work~\cite{tao2020few,li2024mamba,park2024pre,wang2024knowledge}:
\begin{itemize}[leftmargin=*, topsep=2pt, itemsep=0pt]
\item CUB-200: A fine-grained dataset with 200 classes. The base session trains on 100 classes, followed by 10 incremental sessions in a 10-way 5-shot (10 classes with 5 images per class) setting.
\item CIFAR-100 \& miniImageNet: Both datasets contain 100 classes. The base session trains on 60 classes, followed by 8 incremental sessions in a 5-way 5-shot setting.
\end{itemize}
For data augmentation, images are resized to $224 \times 224$ and undergo random resizing, flipping, color jittering, Mixup~\cite{zhang2018mixup}, and Cutout~\cite{devries2017improved}, following methods~\cite{yang2023neural, li2024mamba}.

\vspace{1mm}\noindent \textbf{Evaluation Metrics.} We report the average incremental accuracy across all sessions (\textbf{\textsc{AVG} (\%)}) and the Performance Drop (\textbf{PD} (\%) between the accuracy of the first and the last session, which measures catastrophic forgetting.

\vspace{1mm}\noindent \textbf{Model Architecture.}
We adopt the Vision Transformer (ViT)~\cite{dosovitskiy2020image} as our default backbone due to its reliance on linear projection layers, which aligns well with the design of our low-rank adapter-based parameter-efficient tuning. Following recent SOTA works such as Mamba-FSCIL~\cite{li2024mamba}, we attach a projector that incorporates the Selective State Space Model (SSM)~\cite{gu2023mamba} to process the output features. For classification, we use an ETF-based classifier and optimize the model with the DR loss~\cite{yang2023neural}. Importantly, aside from the loss terms introduced in Mamba-FSCIL for the projector and DR loss, no additional losses are introduced by CKPD-FSCIL.

\vspace{1mm}\noindent \textbf{Backbone Configurations.}
We adopt three types of backbone configurations across experiments:
\begin{itemize}[leftmargin=*, topsep=2pt, itemsep=0pt]
\item ViT-B-CLIP: We use the image encoder from CLIP-ViT-B/16~\cite{radford2021learning} as the default backbone, initialized using \href{https://huggingface.co/openai/clip-vit-base-patch16}{OpenAI’s pre-trained weights}, following CPE-CLIP~\cite{d2023multimodal}, CEC+~\cite{wang2023improved}.
\item ViT-B-IN21K: The ViT-B/16 backbone pre-trained on ImageNet-21K, loaded from the \href{https://storage.googleapis.com/vit_models/augreg/B_16-i21k-300ep-lr_0.001-aug_medium1-wd_0.1-do_0.0-sd_0.0--imagenet2012-steps_20k-lr_0.01-res_224.npz}{official AugReg weights}, following the practice of PriViLege~\cite{park2024pre}.
\item Swin-T-IN1K: We also tested CKPD-FSCIL on the Swin Transformer-Tiny~\cite{liu2021swin} \href{https://download.openmmlab.com/mmclassification/v0/swin-transformer/swin_tiny_224_b16x64_300e_imagenet_20210616_090925-66df6be6.pth}{pretrained on ImageNet-1K} following Comp-FSCIL~\cite{zou2024compositional} and Mamba-FSCIL~\cite{li2024mamba}
\end{itemize}

\vspace{1mm}\noindent \textbf{Training Protocol.} 
All experiments are implemented in PyTorch and run on 8 NVIDIA A100-SXM4 (40GB) GPUs. CKPD-FSCIL is applied only during incremental sessions to adapt the model while preserving its base knowledge.
\begin{itemize}[leftmargin=*, topsep=2pt, itemsep=0pt]
\item Base Session ($t=0$): We use a standard approach by only updating the projector and (optionally) the last transformer block of the ViT-B. All other parameters are frozen to maintain the pre-trained model's generalization ability.
\item Incremental Sessions ($t>0$): Unlike prior methods that keep the backbone entirely frozen~\cite{li2024mamba, yang2023neural}, CKPD-FSCIL selectively fine-tunes low-rank adapters ($B$ and $A$) inserted into the top-$K$ layers chosen by our adaptive layer selection mechanism, alongside the projector. All other weights remain fixed. This enables efficient, non-destructive learning of new tasks. 
\end{itemize}

\vspace{1mm} \noindent \textbf{Input Covariance Calculation.}
Following prior FSCIL work~\cite{tao2020few, liu2022few, peng2022few, park2024pre}, we adopt an \textit{exemplar-based} setting, where only one randomly chosen sample per previously seen class is stored in a small memory buffer. This buffer is used to (1) compute the input covariance matrix $\Sigma_\text{in}$ for our KPD and ALS modules, and (2) rehearse with current-session data to mitigate forgetting. Our key insight is that, even with this standard rehearsal baseline, KPD and ALS provide notable extra gains via intelligent weight reuse and adaptive plasticity allocation. This protocol matches that of the baselines, ensuring fair evaluation.

\vspace{1mm}\noindent \textbf{Hyperparameters.}
For all datasets, the learning rate for adapter modules is 10\% of the projector’s learning rate. The adapter rank $r$ in Eq.\ref{equ:BA} is set to 128 for miniImageNet and CUB-200, and 256 for CIFAR-100. The number of layers selected for adaptation $K$ in Eq.\ref{eq:selected_layers} is 6 for CIFAR-100 and CUB-200, and 9 for miniImageNet.
In experiments without the adaptive layer selection mechanism, we adopt a fixed manual strategy. The ViT-B/16 model contains 12 transformer blocks, each with four linear layers: QKV projection (QKV), post-QKV projection (Proj), and two feed-forward layers (FFN1, FFN2), totaling $N=48$ linear layers. For $K=6$, we select QKV, Proj, and FFN2 from the last two blocks. For $K=9$, the same three layers are selected from the last three blocks. For $K=12$, they are selected from the last four blocks.
\vspace{-1mm}
\begin{itemize}[leftmargin=*, topsep=2pt, itemsep=0pt]
\item CUB-200. Base session uses a batch size of 128, trained for 50 epochs with a initial learning rate of 0.2. Incremental sessions use a batch size of 32, trained for 1000 iterations with a initial learning rate of 0.05.
\item CIFAR-100. Base session uses a batch size of 128 and train for 20 epochs. For the experiments in Tab.~\ref{tab:cifar}, following the PriViLege~\cite{park2024pre} setup, the initial learning rate is set to 2e-4 for ViT-B-IN21K initialization and 1e-5 for ViT-B-CLIP. In other experiments, we use a learning rate of 0.005.
For incremental sessions, we use a batch size of 32. For Tab.~\ref{tab:cifar}, we follow PriViLege's protocol with 20 epochs and a initial learning rate of 5e-5. In other experiments, we train for 1000 iterations using a initial learning rate of 0.25.
\item miniImageNet. The base session uses a batch size of 512, trained for 50 epochs with a initial learning rate of 0.25. Incremental sessions use a batch size of 32, trained for 1000 iterations with a initial learning rate of 0.1.
\end{itemize}

\begin{table*}[t!] 
\renewcommand\arraystretch{1.3}
\begin{center}
\centering
\caption{\textbf{FSCIL performance on CUB-200.}}
\vspace{-2mm}
\resizebox{1\linewidth}{!}{
\begin{tabular}{lllccccccccccccl}
\toprule
\multicolumn{1}{l}{\multirow{2}{*}{\bf Methods}} &\multirow{2}{*}{\bf Venue}&\multirow{2}{*}{\bf Backbone}&\multicolumn{11}{c}{\bf Accuracy in each session (\%)}&\multirow{2}{*}{\bf \textsc{AVG}} &\multirow{2}{*}{\bf PD } \\ 
\cmidrule{4-14}&&&\bf 0&\bf 1&\bf 2&\bf 3&\bf 4&\bf 5&\bf 6&\bf 7&\bf 8&\bf 9&\bf 10& & \\ 
\midrule
Data-free~\cite{liu2022few}&ECCV'22&ResNet-18&75.90&72.14&68.64&63.76&62.58&59.11&57.82&55.89&54.92&53.58&52.39&61.52&23.51\\
MetaFSCIL~\cite{chi2022metafscil}&CVPR'22&ResNet-18&75.90&72.41&68.78&64.78&62.96&59.99&58.30&56.85&54.78&53.82&52.64&61.93&23.26\\
FeSSSS~\cite{ahmad2022few}&CVPR'22&ResNet-18& 79.60&73.46& 70.32&66.38&63.97&59.63&58.19&57.56&55.01&54.31&52.98&62.86&26.62\\
DSN~\cite{yang2022dynamic}&TPAMI'22&ResNet-18&76.06&72.18&69.57&66.68&64.42&62.12&60.16&58.94&56.99&55.10&54.21&63.31&21.85\\
FACT~\cite{zhou2022forward}&CVPR'22&ResNet-18&75.90&73.23&70.84&66.13&65.56&62.15&61.74&59.83&58.41&57.89&56.94&64.42&18.96\\
ALICE~\cite{peng2022few}&ECCV'22&ResNet-18&77.40&72.70&70.60&67.20&65.90&63.40&62.90&61.90&60.50&60.60&60.10&65.75&17.30\\
TEEN~\cite{wang2024few}&NeurIPS'23&ResNet-18&77.26&76.13&72.81&68.16&67.77&64.40&63.25&62.29&61.19&60.32&59.31& 66.63&17.95\\
LIMIT~\cite{zhou2022few}&TPAMI'22&ResNet-18&76.32&74.18&72.68&69.19&68.79&65.64&63.57&62.69&61.47&60.44&58.45&66.67&17.87\\
NC-FSCIL~\cite{yang2023neural}&ICLR'23&ResNet-18&80.45&75.98&72.30&70.28&68.17&65.16&64.43&63.25&60.66&60.01&59.44&67.28&21.01\\
Mamba-FSCIL~\cite{li2024mamba}&Arxiv'24&ResNet-18&80.90&76.26&72.97&70.14&67.83&65.74&65.43&64.12&62.31&62.12&61.65&68.13&19.25\\
\midrule
Finetune &&ViT-B-CLIP& 82.00& 76.72& 70.42& 60.70& 45.24& 25.75& 21.39& 16.84& 13.05& 11.34& 10.39&39.44 &71.61\\
CPE-CLIP~\cite{d2023multimodal}&ICCVW'23&ViT-B-CLIP& 81.58& 78.52& 76.68& 71.86& 71.52& 70.23& 67.66& 66.52& 65.09& 64.47& 64.60&70.79&16.98\\
CEC+~\cite{wang2023improved}&TCSVT'23&ViT-B-CLIP& 82.00& 76.68& 74.97& 72.27& 71.37& 69.89& 68.94& 68.38& 66.89& 67.48& 67.12&71.45&14.88\\
KANet~\cite{wang2024knowledge}&Arxiv'24&ViT-B-CLIP& 82.00& 77.99& 76.68& 74.25& 73.37& 71.55& 70.66& 70.26& 69.13& 69.65& 69.35&73.17&12.65\\
\rowcolor{mygray} CKPD-FSCIL&&ViT-B-CLIP&81.15&78.30&78.51&76.05&74.88&76.12&75.99&76.87&75.47&76.56&75.68&76.87&5.47\\
\rowcolor{mygray} CKPD-FSCIL&&ViT-B-CLIP&87.05&83.11&82.30&79.21&77.93&78.50&78.83&79.47&77.51&78.36&77.68&80.00&9.37\\
\midrule
PL-FSCIL~\cite{tian2024pl}&Arxiv'24&ViT-B-IN1K& 85.16& 85.40& 82.75& 75.22& 77.22& 73.25& 72.39& 70.24& 67.97& 68.33& 69.86&75.25&15.30\\
PriViLege~\cite{park2024pre}&CVPR'24&ViT-B-IN21K& 82.21& 81.25& 80.45& 77.76& 77.78& 75.95& 75.69& 76.00& 75.19& 75.19& 75.08&77.50&7.13\\
ASP-FSCIL~\cite{liu2024few}&ECCV'24&ViT-B-IN1K& 87.10& 86.00& 84.90& 83.40& 83.60& 82.40& 82.60& 83.00& 82.60& 83.00& 83.50&83.83&3.60\\
\rowcolor{mygray} CKPD-FSCIL&&ViT-B-IN21K&88.02&86.38&86.65&85.41&84.96&84.58&85.00&85.34&84.69&85.00&85.33&85.58&2.69\\
\midrule
CLOM~\cite{zou2022margin}& NeurIPS'22&Swin-T-IN1K&86.28&82.85&80.61&77.79&76.34&74.64&73.62&72.82&71.24&71.33&70.50& 76.18&15.78\\
NC-FSCIL~\cite{yang2023neural}& ICLR'23&Swin-T-IN1K&87.53&84.25&81.72&79.10&77.21&75.52&74.51&74.42&72.26&72.86&72.49&77.44&15.04\\
Comp-FSCIL~\cite{zou2024compositional}& ICML'24&Swin-T-IN1K&87.67&84.73&83.03&80.04&77.73&75.52&74.32&74.55&73.35&73.15&72.80&77.90&14.87\\
Mamba-FSCIL&&Swin-T-IN1K&88.13&85.14&83.41& 80.77&77.23&75.73&75.70&75.32&74.18&74.26&74.13&78.55&14.00\\
\rowcolor{mygray} CKPD-FSCIL&&Swin-T-IN1K&88.13&85.81&83.50&80.87&79.02&77.85&77.20&76.50&75.26&75.37&74.77&79.48&13.36\\
\bottomrule
\end{tabular}
}
\label{tab:cub}
\end{center}
\vspace{-4mm}
\end{table*}

\begin{table*}[t!]
\renewcommand\arraystretch{1.1}
\begin{center}
\centering
\caption{\textbf{FSCIL performance on CIFAR-100.} $\ddagger$ indicates results reproduced using their official code.}
\vspace{-2mm}
\resizebox{1\textwidth}{!}{
\begin{tabular}{lllccccccccccl}
\toprule
\multicolumn{1}{l}{\multirow{2}{*}{\bf Methods}} &\multirow{2}{*}{\bf Venue} &\multirow{2}{*}{\bf Backbone}&\multicolumn{9}{c}{\bf Accuracy in each session (\%)}&\multirow{2}{*}{\bf \textsc{AVG}} &\multirow{2}{*}{\bf PD } \\ 
\cmidrule{4-12}& &&\bf 0&\bf 1&\bf 2&\bf 3&\bf 4&\bf 5&\bf 6&\bf 7&\bf 8& & \\
\midrule
DSN~\cite{yang2022dynamic}&TPAMI'22&ResNet-18&73.00&68.83&64.82&62.24&59.16&56.96&54.04&51.57&49.35&60.00&23.65\\
Data-free~\cite{liu2022few}&ECCV'22&ResNet-20&74.40&70.20&66.54&62.51&59.71&56.58&54.52&52.39&50.14&60.78&24.26\\
MetaFSCIL~\cite{chi2022metafscil}&CVPR'22&ResNet-20&74.50&70.10&66.84&62.77&59.48&56.52&54.36&52.56&49.97&60.79&24.53\\
FeSSSS~\cite{ahmad2022few}&CVPR'22&ResNet-20&75.35&70.81&66.70&62.73&59.62&56.45&54.33&52.10&50.23&60.92&25.12\\
C-FSCIL~\cite{hersche2022constrained}&CVPR'22&ResNet-12&77.47&72.40&67.47&63.25&59.84&56.95&54.42&52.47&50.47&61.64&27.00\\
LIMIT~\cite{zhou2022few}&TPAMI'22&ResNet-20&73.81&72.09&67.87&63.89&60.70&57.77&55.67&53.52&51.23&61.84&22.58\\
FACT~\cite{zhou2022forward}&CVPR'22&ResNet-20&74.60&72.09&67.56&63.52&61.38&58.36&56.28&54.24&52.10& 62.24&22.50\\
TEEN~\cite{wang2024few}&NeurIPS'23&ResNet-18&74.92&72.65&68.74&65.01&62.01&59.29&57.90&54.76&52.64& 63.10&22.28\\
ALICE~\cite{peng2022few}&ECCV'22&ResNet-18&79.00&70.50&67.10&63.40&61.20&59.20&58.10&56.30&54.10&63.21&24.90\\
CABD~\cite{zhao2023few}&CVPR'23&ResNet-18&79.45&75.38&71.84&67.95&64.96&61.95&60.16&57.67&55.88&66.14&23.57\\
NC-FSCIL~\cite{yang2023neural}&ICLR'23&ResNet-12&82.52&76.82&73.34&69.68&66.19&62.85&60.96&59.02&56.11&67.50&26.41\\
Mamba-FSCIL~\cite{li2024mamba}&Arxiv '24&ResNet-12&82.80&77.85&73.69&69.67&66.89&63.66&61.48&59.74&57.51&68.14&25.29\\
\midrule
Finetune & &ViT-B-CLIP& 85.67& 81.14& 75.37& 59.68& 50.31& 24.00& 21.03& 16.29& 16.85&47.82 &68.82\\
CEC+~\cite{wang2023improved}&TCSVT'23&ViT-B-CLIP& 85.67& 78.55& 76.51& 73.80& 72.92& 71.67& 71.76& 70.55& 68.90&74.48&16.77\\
KANet~\cite{wang2024knowledge}&Arxiv'24&ViT-B-CLIP& 85.67& 79.94& 78.06& 75.43& 74.43& 73.11& 73.16& 71.95& 70.22&75.77&15.45\\
CPE-CLIP~\cite{d2023multimodal}&ICCVW'23&ViT-B-CLIP& 87.83& 85.86& 84.93& 82.85& 82.64& 82.42& 82.27& 81.44& 80.52&83.42&7.31\\
\rowcolor{mygray} CKPD-FSCIL& &ViT-B-CLIP&81.48&79.92&79.76&76.95&77.40&76.49&76.22&75.71&74.29&77.58&7.19\\
\rowcolor{mygray} CKPD-FSCIL& &ViT-B-CLIP&87.97&86.57&86.23&83.76&84.21&83.69&84.00&83.64&81.84&84.66&6.13\\
\rowcolor{mygray} CKPD-FSCIL& &ViT-B-CLIP&91.42&89.92&89.54&88.04&88.20&87.49&87.60&87.17&85.84&88.36&5.58\\
\midrule
PL-FSCIL~\cite{tian2024pl}&Arxiv'24&ViT-B-IN1K& 89.93& 77.26& 76.12& 68.06& 69.53& 68.21& 70.03& 69.07& 65.73&72.66&24.20\\
PriViLege$^{\ddagger}$~\cite{park2024pre}& CVPR'24&ViT-B-IN21K& 91.57& 89.91& 89.66& 88.21& 88.33& 87.44& 87.59& 87.12& 85.84&88.41&5.73\\
\rowcolor{mygray} CKPD-FSCIL& &ViT-B-IN21K&91.57&90.02&89.81&88.39&88.53&87.69&87.86&87.34&86.21&88.60&5.36\\
\bottomrule
\end{tabular}
}
\label{tab:cifar}
\end{center}
\vspace{-4mm}
\end{table*}

\begin{table*}[t!]
\renewcommand\arraystretch{1.1}
\begin{center}
\centering
\caption{\textbf{FSCIL performance on miniImageNet.}}
\vspace{-2mm}
\resizebox{1\linewidth}{!}{
\begin{tabular}{lllccccccccccl}
\toprule
\multicolumn{1}{l}{\multirow{2}{*}{\bf Methods}} & \multirow{2}{*}{\bf Venue} &\multirow{2}{*}{\bf Backbone}&\multicolumn{9}{c}{\bf Accuracy in each session (\%)}&\multirow{2}{*}{\bf \textsc{AVG}} &\multirow{2}{*}{\bf PD } \\ 
\cmidrule{4-12}& &&\bf 0&\bf 1&\bf 2&\bf 3&\bf 4&\bf 5&\bf 6&\bf 7&\bf 8& & \\ \toprule
DSN~\cite{yang2022dynamic}&TPAMI'22&ResNet-18&68.95&63.46&59.78&55.64&52.85&51.23&48.90&46.78&45.89&54.83&23.06\\
Data-free~\cite{liu2022few}&ECCV'22&ResNet-18&71.84&67.12&63.21&59.77&57.01&53.95&51.55&49.52&48.21&58.02&23.63\\
MetaFSCIL~\cite{chi2022metafscil}&CVPR'22&ResNet-18&72.04&67.94&63.77&60.29&57.58&55.16&52.90&50.79&49.19&58.85&22.85\\
LIMIT~\cite{zhou2022few}&TPAMI'22&ResNet-18&72.32&68.47&64.30&60.78&57.95&55.07&52.70&50.72&49.19&59.06&23.13\\
FACT~\cite{zhou2022forward}&CVPR'22&ResNet-18&72.56&69.63&66.38&62.77&60.60&57.33&54.34&52.16&50.49& 60.70&22.07\\
CABD~\cite{zhao2023few}&CVPR'23&ResNet-18&74.65&70.43&66.29&62.77&60.75&57.24&54.79&53.65&52.22&61.42&22.43\\
TEEN~\cite{wang2024few}&NeurIPS'23&ResNet-18&73.53&70.55&66.37&63.23&60.53&57.95&55.24&53.44&52.08& 61.44&21.45\\
C-FSCIL~\cite{hersche2022constrained}&CVPR'22&ResNet-12&76.40&71.14&66.46&63.29&60.42&57.46&54.78&53.11&51.41&61.61&24.99\\
Regularizer~\cite{akyurek2021subspace}&ICLR'22&ResNet-18&80.37&74.68&69.39&65.51&62.38&59.03&56.36&53.95&51.73&63.71&28.64\\
ALICE~\cite{peng2022few}&ECCV'22&ResNet-12&80.60&70.60&67.40&64.50&62.50&60.00&57.80&56.80&55.70&63.99&24.90\\
SAVC~\cite{song2023learning}&CVPR'23&ResNet-18& 81.12& 76.14& 72.43& 68.92& 66.48& 62.95& 59.92& 58.39& 57.11& 67.05&24.01\\
NC-FSCIL~\cite{yang2023neural}&ICLR'23&ResNet-12&84.02&76.80&72.00&67.83&66.35&64.04&61.46&59.54&58.31&67.82&25.71\\
FeSSSS~\cite{ahmad2022few}&CVPR'22&ResNet-18&81.50&77.04&72.92&69.56&67.27&64.34&62.07&60.55&58.87&68.24&22.63\\
Mamba-FSCIL~\cite{li2024mamba}&Arxiv'24&ResNet-12&84.93&80.02&74.61&71.33&69.15&65.62&62.38&60.93&59.36&69.81&25.57\\
\midrule
CPE-CLIP~\cite{d2023multimodal}&ICCVW'23&ViT-B-CLIP& 90.23& 89.56& 87.42& 86.80& 86.51& 85.08& 83.43& 83.38& 82.77& 86.13&7.46\\
\rowcolor{mygray} CKPD-FSCIL& &ViT-B-CLIP& 90.63&90.52&88.23&87.71&88.06&87.56&85.72&85.62&85.97&87.78&4.66\\
\rowcolor{mygray} CKPD-FSCIL&&ViT-B-CLIP& 96.18&95.38&92.96&91.92&91.25&90.14&88.47&87.93&88.23&91.38&7.95\\
\midrule
PriViLege~\cite{park2024pre}& CVPR'24&ViT-B-IN21K&96.68&96.49&95.65&95.54&95.54&94.91&94.33&94.19&94.10&95.27&2.58\\
\rowcolor{mygray} CKPD-FSCIL& &ViT-B-IN21K&96.4&96.54&95.06&95.43&95.56&95.04&94.36&94.31&94.31&95.22&2.09\\
\bottomrule
\end{tabular}
}
\label{tab:imgnet}
\end{center}
\vspace{-4mm}
\end{table*}

 \vspace{-2mm}\subsection{Comparison with State-of-the-Art Methods}
\label{sec:exps_compare_sota}
To thoroughly assess CKPD-FSCIL, we compare it with a wide range of state-of-the-art (SOTA) methods across three standard benchmarks. As shown in Tabs.~\ref{tab:cub}, \ref{tab:cifar}, and \ref{tab:imgnet}, Across all datasets and backbone initializations, CKPD-FSCIL consistently establishes a new SOTA, demonstrating a superior balance between plasticity and stability. Notably, with a ViT-B-IN21K backbone, our method achieves the \textbf{highest average accuracy} of \textbf{85.58\%} on CUB-200, and \textbf{88.60\%} on CIFAR-100. It also demonstrated the \textbf{lowest performance drop} on three datasets, with a PD of \textbf{2.69\%} on CUB-200, \textbf{5.36\%} on CIFAR-100, and \textbf{2.09\%} on miniImageNet. This superior performance is achieved while adhering to a \textbf{zero-overhead principle} that introduces no additional parameters or inference FLOPs, a critical advantage over most contemporary prompt/adapter tuning methods.

To ensure fairness, we align our base session accuracy with or slightly below strong baselines, confirming that later-session gains stem from our knowledge-preserving decomposition and adaptive layer selection rather than a stronger base session accuracy. Detailed analysis for each benchmark is provided below:

\vspace{1mm}\noindent\textbf{CUB-200 Results.}
On this challenging fine-grained dataset, CKPD-FSCIL demonstrates a particularly strong and consistent advantage across multiple backbone architectures. With a powerful ViT-B-IN21K backbone,
it achieves an average accuracy of 85.58\%, surpassing the second-best method ASP-FSCIL by 1.75\%, while maintaining a performance drop (PD) of only 2.69\%, even improves upon the previous best (ASP-FSCIL at 3.60\%). With ViT-B-CLIP, where we intentionally start with a lower initial accuracy (81.15\%, below KANet’s 82.00\%), our method still achieves 76.87\% average accuracy, outperforming KANet by 3.7\%. This indicates that the improvement stems from a more effective continual learning mechanism rather than an initial performance advantage. On the Swin-T-IN1K backbone, we likewise obtain the highest average accuracy (79.48\%) and the lowest PD (13.36\%), verifying the architecture-agnostic nature of our approach.

\noindent\textbf{CIFAR-100 Results.}
On CIFAR-100, CKPD-FSCIL achieves SOTA performance with both high accuracy and minimal forgetting. Using ViT-B-IN21K, it reaches an average accuracy of 88.60\% and the lowest PD of 5.36\%, slightly outperforming PriViLege (88.41\%, 5.73\%) while avoiding extra parameters or inference costs. Even when starting with a similar base accuracy to strong baselines on ViT-B-CLIP (87.97\% vs. CPE-CLIP’s 87.83\%), CKPD-FSCIL maintains a clear lead in later sessions (84.66\% vs. 83.42\% \textsc{AVG}), confirming that gains stem from more effective continual adaptation rather than initial advantage. These results demonstrate that our framework delivers top-tier incremental learning without the growing complexity common in prior methods.

\vspace{1mm}\noindent\textbf{miniImageNet Results.}
On the competitive miniImageNet, CKPD-FSCIL achieves SOTA accuracy while achieving superior long-term stability across multiple backbones under fair starting conditions. With ViT-B-IN21K, we match PriViLege at the base (96.40\% vs. 96.68\%) and obtain comparable \textsc{AVG} (95.22\% vs. 95.27\%), while achieving the lowest PD among all methods (2.09\%), indicating stronger resistance to catastrophic forgetting. On ViT-B-CLIP, starting from a comparable base (90.63\% vs. CPE-CLIP’s 90.23\%), our method delivers higher average accuracy (87.78\% vs. 86.13\%) and lower PD (4.66\% vs. 7.46\%). 

\vspace{-2mm}\subsection{Ablation Studies}
\label{sec:exps_ablation}
We conduct a series of ablation studies to validate the effectiveness and robustness of our key contributions.

\subsubsection{Effectiveness of KPD}
\label{sec:ablation_kpd_effectiveness}
We compare our covariance-guided KPD against several representative adaptation strategies, including SVD (PiSSA)~\cite{meng2024pissa}, ASVD~\cite{yuan2023asvd}, LoRA~\cite{hu2022lora}, full fine-tuning (Full Adapt), and full backbone freezing (Freeze). All methods adopt the unified adapter formulation $W = W_{\text{frozen}} + BA$, where $W_{\text{frozen}}$ is the frozen weight component and $BA$ is the low-rank trainable adapter, differing in how $W_{\text{frozen}}$, $B$, and $A$ are derived:
\begin{itemize}[leftmargin=*, topsep=2pt, itemsep=0pt]
\item \textbf{Freeze}: No adaptation ($BA=0$); all weights frozen.
\item \textbf{Full Adapt}:All weights trainable ($W_{\text{frozen}}=0$).
\item \textbf{LoRA}: Adapters $B$ and $A$ are randomly initialized; $W_{\text{frozen}} = W$.
\item \textbf{SVD} (PiSSA)~\cite{meng2024pissa}: Performs SVD on $W$; top $R{-}r$ singular components form $W_{\text{frozen}}$, bottom $r$ form $BA$.
\item \textbf{ASVD~\cite{yuan2023asvd}}: Applies SVD to $WA$, where $A$ is the activation matrix; then recovers $W$ via $U\Sigma V^\top \gF^{-1}$; top $R{-}r$ singular components form $W_{\text{frozen}}$, bottom $r$ form $BA$.
\item \textbf{CKPD-FSCIL}: Applies SVD to the class-covariance-weighted matrix $W\Sigma_x$; then reconstructs $W$ via $U\Sigma V^\top \Sigma^{-1}$; top $R{-}r$ singular components form $W_{\text{frozen}}$, bottom $r$ form $BA$.
\end{itemize}

For fair comparison, all methods use the same base session training settings with same base session accuracy. We disable adaptive layer selection and insert adapters into the same $K$ layers in the last transformer blocks ($K{=}12$ for CIFAR-100, $K{=}6$ for CUB-200), using the same adapter rank ($r{=}256$ and $r{=}128$, respectively). Experiments use ViT-B-CLIP as the pretrained backbone.

\begin{table}[t!]
\centering
\caption{\textbf{Comparison of adaptation strategies on CIFAR-100 and CUB-200.} \textsc{Base (0)} denotes the base class accuracy after the base (0) session. \textsc{Base (0)} or \textsc{Base (10)} denotes the base class accuracy after the final incremental session.}
\vspace{-2mm}
\label{tab:peft_comparison}
\resizebox{\linewidth}{!}{
\begin{tabular}{l|ccc|ccc}
\toprule
\multirow{2}{*}{\textbf{Methods}} &  \multicolumn{3}{c}{\textbf{CIFAR-100}}& \multicolumn{3}{c}{\textbf{CUB-200}}\\
&  \textbf{\textsc{Base (0)}} &\textbf{\textsc{Base (8)}}  & \textbf{\textsc{AVG}}& \textbf{\textsc{Base} (0)} &\textbf{\textsc{Base (10)}} & \textbf{\textsc{AVG}}\\
\midrule
Freeze& 86.77&79.63  & 74.76 & 87.05&82.93 & 77.77 \\
Full Adapt&  86.77&77.80  & 73.14 & 87.05&77.97 & 73.19 \\
LoRA~\cite{hu2022lora} &  86.77&78.93  & 75.21 & 87.05&81.63 & 78.35 \\
SVD &  86.77&78.85  & 75.42 & 87.05&82.86 & 78.44 \\
ASVD~\cite{yuan2023asvd} &  86.77&75.82  & 74.11 & 87.05&81.08 & 77.63 \\
\rowcolor{mygray} CKPD-FSCIL&  86.77&80.65  & 76.01 & 87.05&83.66 & 79.21 \\
\bottomrule
\end{tabular}
}
\vspace{-4mm}
\end{table}

Results in Tab.~\ref{tab:peft_comparison} show that CKPD-FSCIL achieves highest average and final base class accuracy. Unlike SVD or ASVD, which lack a principled way to align decomposition with task-relevant knowledge, CKPD leverages input covariance to accurately separate redundant parameters from essential ones. This leads to better adaptability while preserving stability, outperforming other methods in both accuracy and resistance to forgetting.

\subsubsection{Effectiveness of Knowledge Separation}
\label{sec:ablation_kpd_dropout}
\begin{table}[t!]
\centering
\caption{\textbf{Robustness to adapter dropout.} CIFAR-100 performance after session 3 under increasing dropout rates applied to the learnable adapter’s output. \textsc{Base}: accuracy on all base classes. \textsc{Novel}: accuracy on all incrementally introduced classes.}
\label{tab:dropout_session3}
\resizebox{0.8\linewidth}{!}{
\begin{tabular}{l|cc|cc}
\toprule
\multirow{2}{*}{\textbf{Methods}} & \multicolumn{2}{c|}{\textbf{Dropout 0.6}} & \multicolumn{2}{c}{\textbf{Dropout 0.7}} \\
& \textbf{\textsc{Base}} & \textbf{\textsc{Novel}} & \textbf{\textsc{Base}} & \textbf{\textsc{Novel}} \\
\midrule
SVD & 71.53 & 56.90 & 55.32 & 51.40 \\
ASVD~\cite{yuan2023asvd} & 73.13 & 58.70 & 56.87 & 51.90 \\
\rowcolor{mygray} CKPD-FSCIL & 74.63 & 59.30 & 63.07 & 55.00 \\
\bottomrule
\end{tabular}
}
\end{table}

A superior decomposition should not only be effective but also robust, meaning it should concentrate essential knowledge cleanly into the frozen part, making the model's performance less sensitive to perturbations in the adaptable part. To further probe the quality of the decomposition, we conduct an adapter dropout experiment using a ViT-B-CLIP backbone. This test cleverly simulates perturbations to the adaptable part, directly quantifying the ``purity" of our knowledge-redundancy separation. After training up to session 3, we apply increasingly high dropout rates to the output of the learnable adapter ($W_{\text{learnable}} \gF_\text{in}$) during evaluation. All methods started with the same pre-dropout accuracy to ensure a fair comparison.

Tab.~\ref{tab:dropout_session3} shows that CKPD maintains the highest accuracy on both base and novel classes under heavy dropout. For instance, at a dropout rate of 0.7, our method retains 63.07\% base accuracy, significantly outperforming SVD (55.32\%) and ASVD (56.87\%). This result provides strong evidence that our covariance-guided approach achieves a cleaner and more robust separation of knowledge. It effectively isolates critical information in the stable, frozen subspace, leaving the adapter with truly redundant capacity that is less critical for the model's core performance.

\subsubsection{Effectiveness of Continuous Decomposition (CKPD)}
\label{sec:ablation_ckpd_effectiveness}

\begin{table}[t!]
\centering
\caption{\textbf{FSCIL Performance comparison of KPD and CKPD on CIFAR-100 and CUB-200 dataset.}}
\label{tab:kpd_ckpd}
\subfloat{
\resizebox{\linewidth}{!}{
\begin{tabularx}{\linewidth}{l*{11}{>{\centering\arraybackslash}X}}
\toprule
\multirow{2}{*}{\textbf{Mode}} & \multicolumn{9}{c}{\textbf{CIFAR-100 Accuracy in each session (\%)}} & \scriptsize  \multirow{2}{*}{\textbf{AVG}} \\
\cmidrule{2-10}
& \bf 0 & \bf 1 & \bf 2 & \bf 3 & \bf 4 & \bf 5 & \bf 6 & \bf 7 & \bf 8 & \\
\midrule
KPD& 86.7& 82.4& 79.6& 73.2& 68.2& 61.8& 59.8& 56.3& 54.4& 69.2\\
\rowcolor{mygray}CKPD & 86.7& 82.4& 79.8& 74.7& 70.0& 65.5& 63.8& 60.8& 58.2& 71.3\\
\bottomrule
\end{tabularx}
}
}

\subfloat{
\resizebox{\linewidth}{!}{
\begin{tabularx}{\linewidth}{l*{12}{>{\centering\arraybackslash}X}}
\toprule
\multirow{2}{*}{\textbf{Mode}} & \multicolumn{11}{c}{\textbf{CUB-200 Accuracy in each session (\%)}} & \scriptsize  \multirow{2}{*}{\textbf{AVG}} \\
\cmidrule{2-12}
& \bf 0 & \bf 1 & \bf 2 & \bf 3 & \bf 4 & \bf 5 & \bf 6 & \bf 7 & \bf 8 & \bf 9 & \bf 10 \\
\midrule
KPD& \scriptsize  87.1& \scriptsize 82.3& \scriptsize 81.3& \scriptsize 77.6& \scriptsize 72.8& \scriptsize 73.6& \scriptsize 73.4& \scriptsize 73.5& \scriptsize 71.9& \scriptsize 72.1& \scriptsize 70.9& 76.0\\
\rowcolor{mygray}CKPD & \scriptsize 87.1& \scriptsize 82.3& \scriptsize 81.6& \scriptsize 78.5& \scriptsize 75.3& \scriptsize 75.5& \scriptsize 76.0& \scriptsize 76.4& \scriptsize 75.1& \scriptsize 75.4& \scriptsize 74.5& 78.0\\
\bottomrule
\end{tabularx}
}
}
\vspace{-2mm}
\end{table}

\begin{figure}[t!]
\centering
\includegraphics[width=\linewidth]{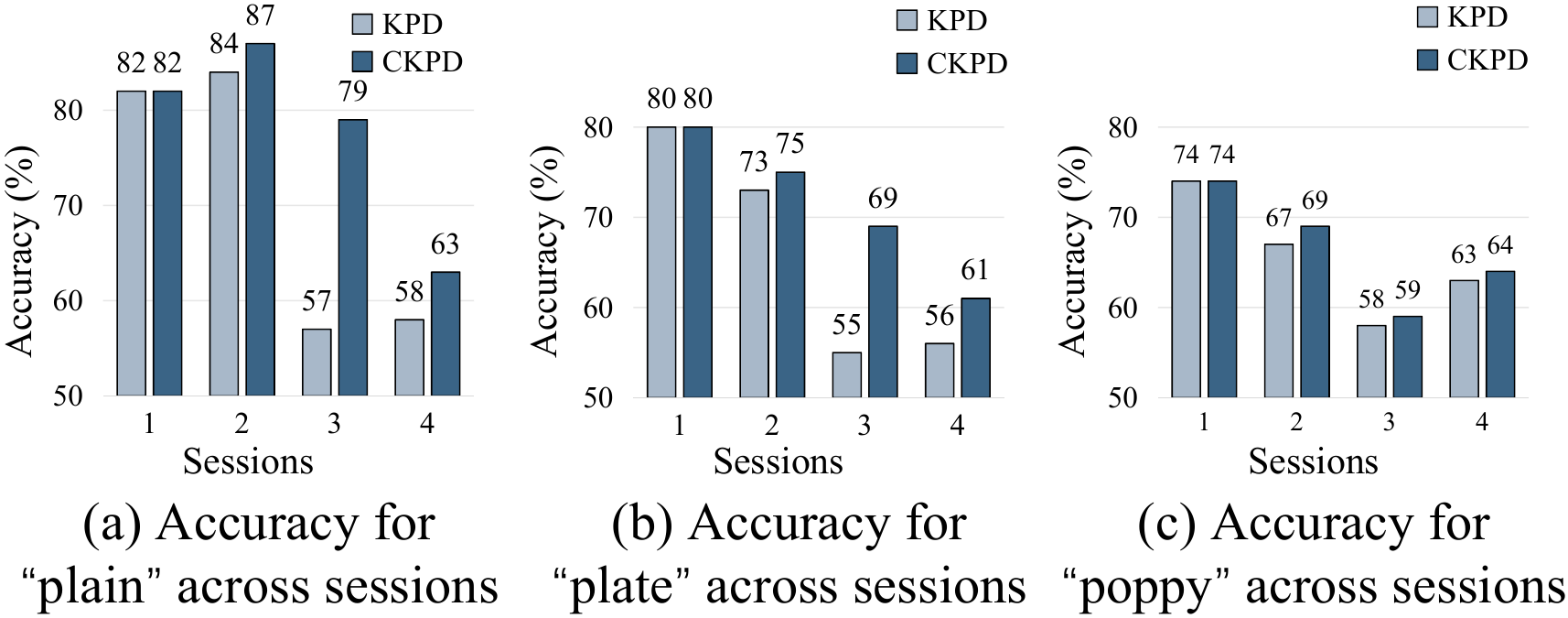}
\vspace{-4mm}
\caption{\textbf{Class-wise accuracy across incremental sessions for three CIFAR-100 novel classes introduced in session~1}: (a) “plain,” (b) “plate,” and (c) “poppy”. The \textit{x}-axis denotes the session index, and the \textit{y}-axis shows the classification accuracy (\%) for each class at the corresponding session.}
\label{fig:ckpd_vs_kpd}
\vspace{-4mm}
\end{figure}
To evaluate the impact of making decomposition continuous, we compare CKPD, where the KPD is recalibrated at each session, with the static KPD baseline that performs decomposition only once after the base session. This isolates the benefit of adapting the knowledge–redundancy partition to the model’s evolving feature space. By dynamically updating the covariance statistics and subspace split, CKPD is expected to better preserve prior knowledge, maintain fine-grained discrimination, and mitigate subspace obsolescence over long sequences of incremental tasks.

\vspace{1mm}\noindent \textbf{Performance Comparison.}
As shown in Table \ref{tab:kpd_ckpd}, CKPD consistently surpasses static KPD on both CIFAR-100 and CUB-200. On CIFAR-100, CKPD improves average accuracy by +2.18\% (71.33\% vs. 69.15\%), while on CUB-200 it gains +1.96\% (77.98\% vs. 76.02\%). These improvements highlight the importance of continuously updating the decomposition to reflect the model’s most recent knowledge base, enabling better alignment between preserved subspaces and evolving task requirements.

\vspace{1mm}\noindent \textbf{Class-wise Knowledge Retention Analysis.}
We further analyze retention at the individual class level by tracking accuracy over time for three novel CIFAR-100 classes introduced in session 1: plain, plate, and poppy (Fig.~\ref{fig:ckpd_vs_kpd}).
In session~1, KPD and CKPD achieve similar accuracy. 
Both methods start with similar accuracy in session 1, but from session 2 onward, CKPD consistently outperforms KPD across all three classes, with gains up to $+22\%$ (e.g., “plate” in session~3). For plate, KPD’s accuracy drops by 25\% between sessions 1 and 3, whereas CKPD limits the drop to 11\%, demonstrating CKPD’s ability to preserve fine-grained, discriminative features without erasing prior knowledge that static decompositions suffer over time.

\subsubsection{Effectiveness of Adaptive Layer Selection}
\label{sec:ablation_cals_effectiveness}

\begin{figure}[t!]
\centering
\includegraphics[width=1\linewidth]{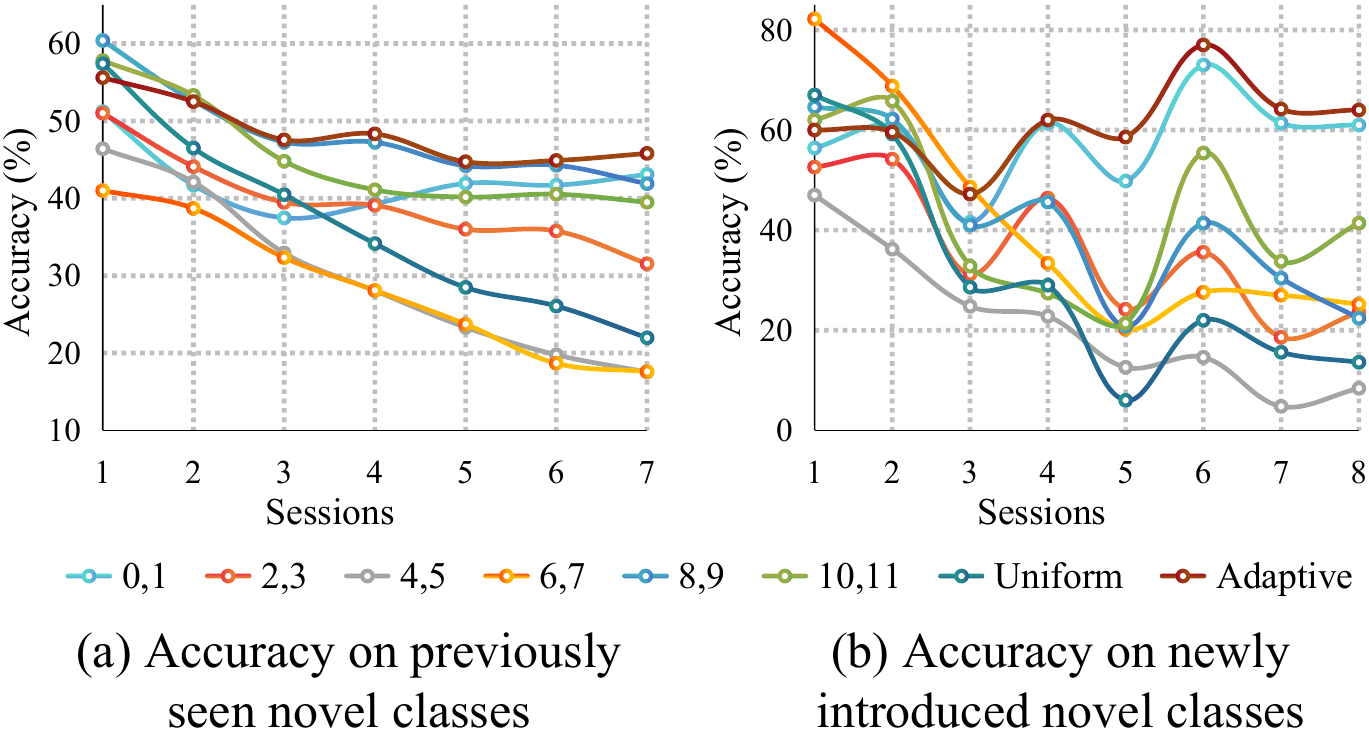}
\vspace{-2mm}
\caption{\textbf{Comparison of adaptive vs. manual layer selection on CIFAR-100.} (a) Average accuracy on all previously seen novel classes (measuring knowledge retention). (b) Accuracy on the newly introduced classes in the current session (measuring plasticity).}
\label{fig:ada_select}
\vspace{-4mm}
\end{figure}

\begin{figure}[t!]
\centering
\includegraphics[width=1\linewidth]{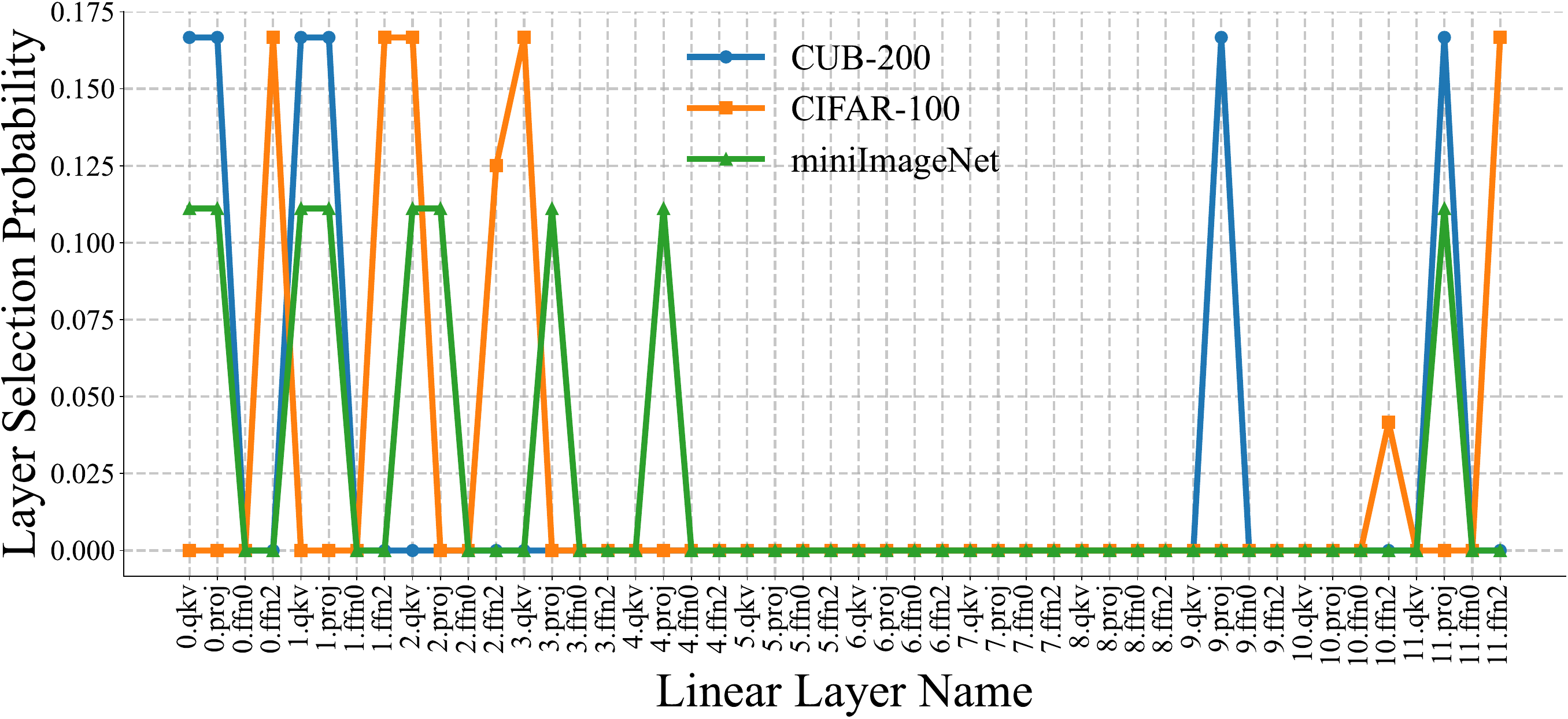}
\vspace{-6mm}
\caption{\textbf{Layer selection frequency across sessions on three datasets.}
 The x-axis lists all linear layers in the ViT backbone, including QKV projections (\texttt{qkv}), attention outputs (\texttt{proj}), and feed-forward layers (\texttt{ffn0}, \texttt{ffn1}), indexed by block (e.g., \texttt{0.proj} refers to the attention output projection in block 0). The y-axis indicates the selection probability under our adaptive strategy.}
\label{fig:linear-selection}
\vspace{-4mm}
\end{figure}
We demonstrate the superiority of our ASR-based selection by first showing its performance advantage and then revealing its layer selection patterns.

\noindent \textbf{Performance Comparison of Adaptive vs. Manual Layer Selection}
We compare our ASR-guided Adaptive Layer Selection (ALS) with two common placement strategies on CIFAR-100 using a ViT-B-CLIP backbone.
All methods adapt $K{=}6$ linear layers per session:
\begin{itemize}[leftmargin=*, topsep=2pt, itemsep=0pt]
\item \textbf{Manual ($i,j$)}: Adapts three linear modules (QKV projection, output projection, and the second linear layer of FFN) in two manually selected transformer blocks. We examine block pairs from shallow (0,1), middle (4,5), to deep (10,11).
\item \textbf{Uniform}: Places adapters in second FFN layer of blocks \{0, 2, 4, 6, 8, 10\} to distribute adaptation capacity across depth.
\item \textbf{Adaptive (Ours)}: Selects the 6 linear layers with the lowest ASR scores across all 12 blocks, dynamically identifying the most adaptable positions each session.
\end{itemize}

We evaluate each strategy on two metrics: knowledge retention (accuracy on previously seen novel classes) and plasticity (accuracy on newly introduced classes).
As shown in Fig.~\ref{fig:ada_select}, ALS (red line) consistently achieves the best performance on both metrics. For \textit{knowledge retention} (Fig.~\ref{fig:ada_select}~(a)), our Adaptive strategy maintains the highest accuracy across sessions, indicating strong resistance to forgetting. For \textit{plasticity} (Fig.~\ref{fig:ada_select}~(b)), ALS again ranks among the top performers. In contrast, the manual strategies reveal a clear trade-off: middle-block adaptation (e.g., blocks {4,5} or {6,7}) causes severe forgetting, while late-block adaptation ({10,11}) better preserves prior knowledge but hinders new task learning.

\vspace{1mm}\noindent \textbf{Layer Selection Patterns Analysis}
To further assess the behavior of our adaptive strategy, we analyze the selection frequency of each linear layer across incremental sessions on the three datasets (Fig.\ref{fig:linear-selection}). ALS exhibits a clear and consistent trend: it largely bypasses the middle blocks (4–8) and instead favors adaptation in the early (0–3) and late blocks (9–11).
This “middle-layer freezing” behavior aligns with the observations in Fig.\ref{fig:ada_select}: middle layers encode core, general-purpose representations that are highly sensitive to changes, while early and late layers contain more redundant capacity that can be repurposed with minimal impact on prior knowledge.
Without relying on manually designed heuristics, ALS automatically avoids these high-risk layers and focuses on the network’s more plastic periphery. This principled plasticity allocation, which respects the hierarchical structure of the Vision Transformer, contributes to more stable and effective continual learning.

\subsubsection{Impact of Hyperparameters}
\label{sec:ablation_hyperparames}
\begin{figure}[t]
\centering
\subfloat[\small{\textrm{(a) Impact of adapter rank $r$}}]{
\includegraphics[width=0.47\linewidth]{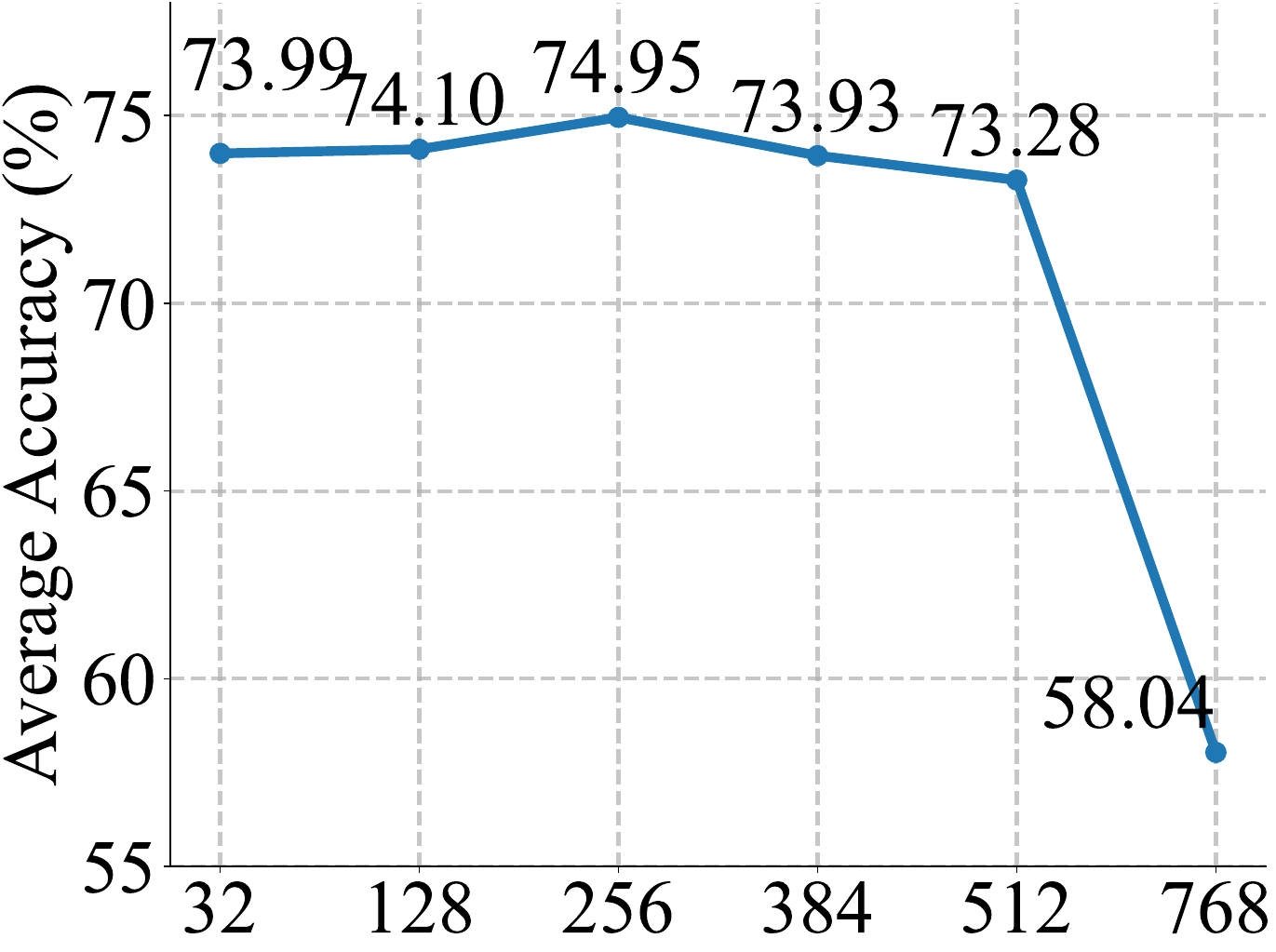}
}
\hfill
\subfloat[\small{\textrm{(b) Impact of adaptable layers $K$}}]{
\includegraphics[width=0.47\linewidth]{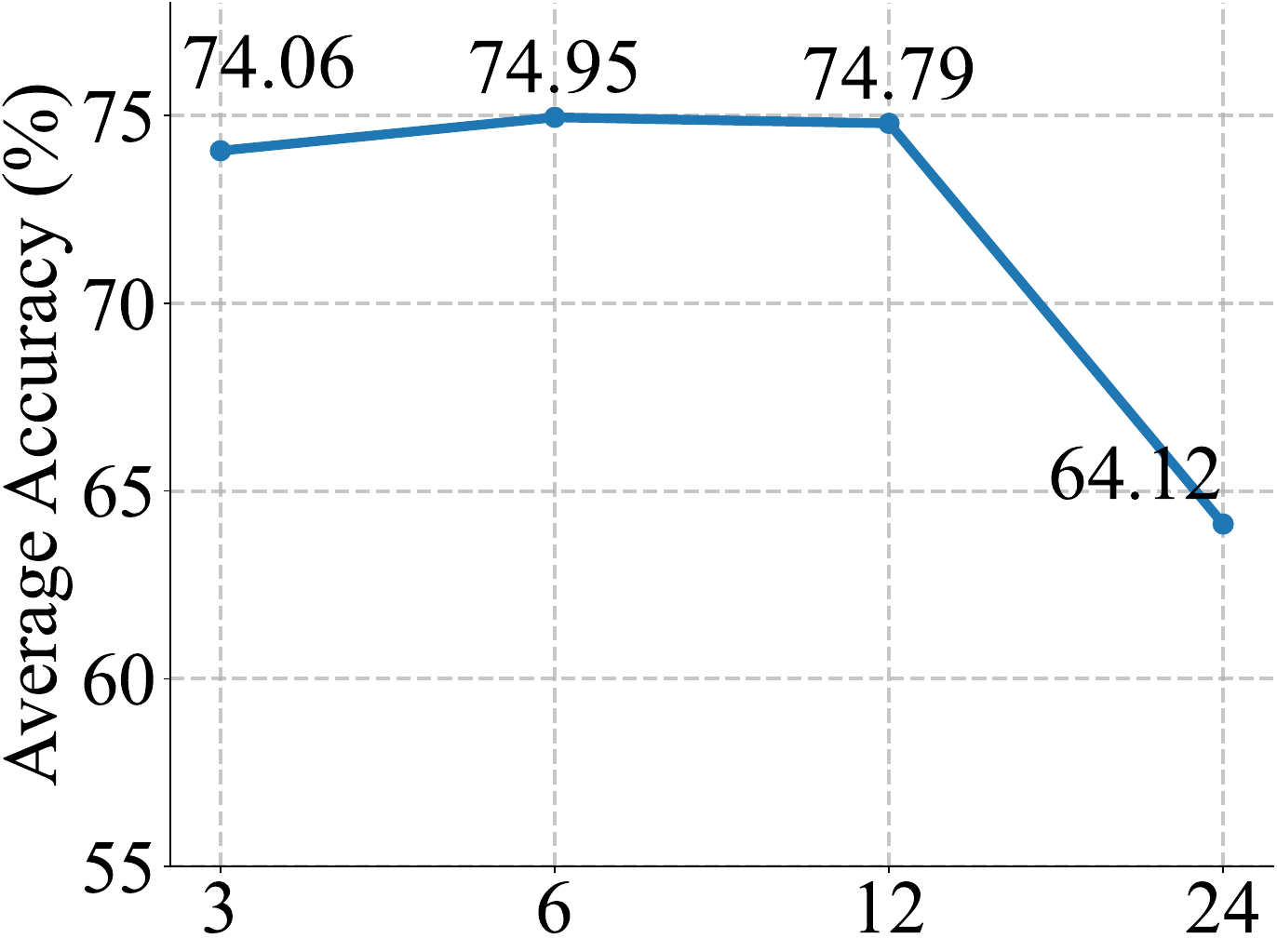}
}
\caption{\textbf{Impact of hyperparameters on CKPD-FSCIL performance.} 
(a) Adapter rank $r$ controls the capacity of each adapted layer. 
(b) The number of adaptable layers $K$ controls the overall adaptation capacity across the network.}
\label{fig:ablation_r_k}
\end{figure}

We further evaluate CKPD-FSCIL’s robustness to two key hyperparameters on CIFAR-100 with ViT-B-CLIP: adapter rank $r$, which controls the \emph{capacity} of each adapted layer, and the number of adaptable layers $K$, which determines how many layers are adapted and thus the overall capacity of the network. 
The results show that our method is robust to a wide range of these hyperparameters.

\vspace{1mm}\noindent \textbf{Impact of adapter rank $r$.}
As shown in Fig.\ref{fig:ablation_r_k}~(a), CKPD-FSCIL maintains competitive performance across a wide range ($32 \leq r \leq 512$), with the best result at $r=256$. Even with small ranks (e.g., $r=32$), CKPD-FSCIL achieves $73.99\%$, indicating strong adaptation ability under highly compact parameterization. Notably, as $r$ increases to cover the full subspace (e.g., $r=768$), performance drops significantly. This validates our core assumption that constraining updates to a low-rank redundant subspace is critical; excessive plasticity can be as detrimental as too little, leading to forgetting.

\vspace{1mm}\noindent \textbf{Impact of number of adaptable layers $K$.}
As shown in Fig.~\ref{fig:ablation_r_k}~(b), the model achieves stable performance when adapting a moderate number of layers ($3 \leq K \leq 12$ out of 48 linear layers in ViT-B). The peak is observed at $K=6$, while $K=12$ only shows a negligible drop. However, when adapting half of all linear layers ($K=24$), performance drops sharply to 64.12\%, suggesting that excessive adaptation undermines the preserved knowledge subspace.

\vspace{-2mm}\subsection{Generalization and Robustness Analysis}
\label{sec:exps_generalization}

\subsubsection{Scalability to Large-Scale Datasets}
\label{sec:exps_generalization_in1k}
\begin{table}[t!]
\centering
\caption{\textbf{FSCIL performance on ImageNet-1K.}}
\label{tab:imagenet1k-sessions}
\resizebox{\linewidth}{!}{
\begin{tabularx}{\linewidth}{l*{13}{>{\centering\arraybackslash}X}}
\toprule
\scriptsize  \multirow{2}{*}{\bf Methods} & \multicolumn{11}{c}{\bf Accuracy in each session (\%)} & \multirow{2}{*}{\textbf{\textsc{Avg}}} \\ 
\cmidrule(lr){2-12}
& \bf \scriptsize 0 & \bf \scriptsize 1 & \bf \scriptsize 2 & \bf \scriptsize 3 & \bf \scriptsize 4 & \bf \scriptsize 5 & \bf \scriptsize 6 & \bf \scriptsize 7 & \bf \scriptsize 8 & \bf \scriptsize 9 & \bf \scriptsize 10 & \\ 
\midrule
\scriptsize  Frozen & \scriptsize 79.3& \scriptsize 75.8& \scriptsize 73.3& \scriptsize 71.0& \scriptsize 68.8& \scriptsize 67.4& \scriptsize 65.6& \scriptsize 64.2& \scriptsize 63.0& \scriptsize 61.7& \scriptsize 60.2& \scriptsize 68.2\\
\scriptsize  Finetune & \scriptsize 79.3& \scriptsize 75.3& \scriptsize 73.0& \scriptsize 71.3& \scriptsize 69.2& \scriptsize 68.1& \scriptsize 66.3& \scriptsize 64.9& \scriptsize 63.6& \scriptsize 62.4& \scriptsize 61.0& \scriptsize 68.6\\
\rowcolor{mygray} \tiny CKPD-FSCIL & \scriptsize 79.3& \scriptsize 75.8& \scriptsize 74.1& \scriptsize 72.3& \scriptsize 70.2& \scriptsize 68.9& \scriptsize 67.1& \scriptsize 65.6& \scriptsize 64.3& \scriptsize 63.2& \scriptsize 61.5& \scriptsize 69.3\\
\bottomrule
\end{tabularx}
}
\vspace{-4mm}
\end{table}
We test CKPD-FSCIL on a challenging large-scale ImageNet-1K benchmark~\cite{deng2009imagenet}, which presents a significant challenge due to its large number of classes and long incremental sequence. We configure the experiment with a base session of 500 classes, followed by 10 incremental sessions. Each incremental session introduces 50 new classes in a 5-shot setting, and training is performed for 1000 iterations per session.
We compare with two strong baselines, all initialized from the same CLIP-pretrained ViT-B/16: \textbf{1) Frozen}: backbone frozen, only projector and classifier trained. \textbf{2) Finetune}: fine-tunes the final two blocks of the ViT, which is a common heuristic for balancing stability and plasticity.

Tab.~\ref{tab:imagenet1k-sessions} shows that CKPD-FSCIL achieves the highest accuracy in all sessions and the best average accuracy of 69.28\%. The performance gap between our method and the baselines widens over time, indicating a stronger long-term balance between stability and plasticity. This confirms CKPD-FSCIL's robustness and scalability for realistic large-scale continual learning scenarios.

\subsubsection{Generalization to Class-Incremental Learning (CIL)}
\label{sec:exps_generalization_cil}

\begin{table}[t!]
\centering
\caption{\textbf{Generalization to CIL on CUB-200.} 
Replacing the additive adapters in EASE with our repurposing-based adapters yields higher accuracy.}
\label{tab:ease}
\resizebox{\linewidth}{!}{
\begin{tabular}{l S[table-format=2.2] S[table-format=2.2] S[table-format=2.2]}
\toprule
\textbf{Method} & {\textbf{CODA-Prompt~\cite{smith2023coda}}} & {\textbf{EASE~\cite{zhou2024expandable}}} & {\textbf{CKPD-FSCIL}} \\
\midrule
Avg. Acc.   & 84.00 & 92.23 & \cellcolor[HTML]{E9F1F6}92.42 \\
Final Acc.  & 73.37 & 86.81 & \cellcolor[HTML]{E9F1F6}87.23 \\
\bottomrule
\end{tabular}
}
\end{table}

To demonstrate the versatility of our principles beyond the few-shot setting, we integrate our modules into the SOTA CIL method~\cite{zhou2024expandable}, replacing its randomly initialized additive adapters with adapters constructed via our KPD to reuse redundant weights, and allocating layers via CALS to target layers with the highest and safest adaptation potential. Following EASE’s protocol, we evaluate on the CUB-200 dataset with 0 base classes and 10 incremental sessions (10 new classes per session, abundant data; B100-Inc10), using a ViT-B/16 pretrained on ImageNet-21K.

As shown in Tab.~\ref{tab:ease}, our method surpasses both EASE and the strong prompt-based baseline CODA-prompt, achieving the highest average accuracy (92.42\%) and best final-session accuracy (87.23\%).
This demonstrates that reusing redundant parameters through CKPD and CALS enables more effective adaptation than adding new modules, even with abundant training data, underscoring the broad applicability of our approach to continual learning.

\subsubsection{Robustness to Representative Sample Selection}
\label{sec:exps_generalization_random_seeds}

\begin{table}[t!]
\centering
\caption{\textbf{Performance stability across different random seeds.} Results are average accuracy (\%) from five independent runs. The low standard deviation demonstrates robustness to the random selection of representative exemplars.}
\resizebox{\linewidth}{!}{
\begin{tabular}{lcccccc}
\toprule
\textbf{Dataset} & \textbf{Run 1} & \textbf{Run 2} & \textbf{Run 3} & \textbf{Run 4} & \textbf{Run 5} & \textbf{Mean$\pm$ Std} \\ 
\midrule
CUB-200 & 79.00& 79.19& 79.04& 79.06& 79.28& 79.11$\pm$ 0.10\\ 
miniImageNet & 90.66& 90.41& 90.75& 90.51& 90.65& 90.60$\pm$ 0.12\\
\bottomrule
\end{tabular}}
\label{tab:seeds}

\end{table}
Our KPD module relies on a small set of past-class samples (one per class) to model the structure of prior knowledge. For data selection strategy, we randomly select one sample per class following prior works~\cite{wu2019large,kukleva2021generalized}.
To assess sensitivity to sample choice, we conduct five independent runs of our entire framework on CUB-200 and miniImageNet. For each run, we use a different random seed, which affects both the model initialization and, crucially, the random selection of exemplars for the representative buffer at each session.

As shown in Tab.~\ref{tab:seeds}, the standard deviations are minimal ($\pm$0.10 on CUB-200 and$\pm$0.12 on miniImageNet), confirming that the covariance structure of each class is well captured even by a single random sample. This validates that our simple random selection is sufficient for stable, SOTA performance.

\subsubsection{Efficiency Analysis}
\label{sec:exps_efficiency}
A key strength of CKPD-FSCIL is its efficiency, delivering SOTA performance with zero inference overhead and minimal training cost, making it practical for long-term continual learning.

\begin{table}[t!]
\centering
\caption{\textbf{Inference cost comparison.} FLOPs (G) and parameters (M) for the initial model (Init) and the final incremental model (Final Inc.). Red arrows indicate growth relative to the initial model.}
\vspace{-2mm}
\resizebox{\linewidth}{!}{
\begin{tabular}{l|cc|cc}
\toprule
\multirow{2}{*}{\textbf{Methods}} & \multicolumn{2}{c|}{\textbf{FLOPs (G)}} & \multicolumn{2}{c}{\textbf{Parameters (M)}} \\
& \textbf{Init} & \textbf{Final Inc.} & \textbf{Init} & \textbf{Final Inc.} \\
\midrule
ASP-FSCIL~\cite{liu2024few} & 85.80& 173.92 \color{red}{ 102.7\%}& 17.58& 35.74 \color{red}{ 103.3\%}\\
\rowcolor{mygray} CKPD-FSCIL & 85.80& 85.80 \quad (=) & 17.58& 17.58 \quad (=)\\
\bottomrule
\end{tabular}
}
\label{tab:flops_params_comparison}
\end{table}

\vspace{1mm}\noindent\textbf{Inference Efficiency: Zero-Overhead Adaptation.}
Unlike additive methods that accumulate new parameters with each session, our framework adheres to a strict zero-overhead principle. This is a direct result of our select-decompose-train-merge cycle, where the final merge step fully integrates all learned changes back into the original weight structure, leaving no extra modules.
As demonstrated in Tab.~\ref{tab:flops_params_comparison}, this design keeps both the parameter count and inference FLOPs \textbf{constant} across all incremental sessions, while methods like ASP-FSCIL more than double their compute and memory costs. This weight-repurposing design ensures scalability without sacrificing accuracy.

\begin{table}[t!]
\centering
\caption{\textbf{Training time analysis on CUB-200.} Wall-clock time (hours) across 10 incremental sessions. Our offline steps constitute only a small fraction of total training time.}
\vspace{-2mm}
\label{tab:training_time}
\resizebox{\linewidth}{!}{
\begin{tabular}{lcc}
\toprule
\multirow{2}{*}{\textbf{Component}} & \textbf{Total Time} & \textbf{Percentage} \\
&(\textbf{Hours})& \textbf{(\%)} \\
\midrule
\textbf{Offline Overhead (Our Contribution)}: & \\
\quad - Input Covariance Calculation & 0.35 &\\
\quad - Knowledge-Preserving Decomposition & 0.19 &\\
\quad - ASR-guided Adaptive Layer Selection & 0.04 &\\
\rowcolor{mygray} \quad \textbf{Total Offline Overhead} & \textbf{0.58} & \textbf{8.4} \\
\midrule
\textbf{Online Training (Backpropagation)} & \textbf{6.30} & \textbf{91.6} \\
\midrule
\textbf{Total Training Time} & \textbf{6.88} & \textbf{100} \\
\bottomrule
\end{tabular}
}
\end{table}

\vspace{1mm}\noindent\textbf{Training Efficiency: Minimal Offline Overhead.}
Our framework introduces several analytical steps (input covariance calculation, KPD, and ASR-guided layer selection) that are performed offline before each training session. To quantify their cost, we measure the wall-clock time for each component of our method across 10 incremental sessions on CUB-200, with each session trained for 2000 iterations. We used a ViT-B-CLIP on a single NVIDIA A100 GPU.
As shown in Tab.~\ref{tab:training_time}, these steps take just 0.58 hours in total across 10 sessions, a minimal 8.4\% of the total training duration, with 91.6\% spent on the standard backpropagation training.
This minimal overhead is attributable to two factors: first, the computations are guided by an lightweight representative buffer (one sample per past class). Second, these steps are performed only once per session, outside the main training loop. 

This analysis confirms that CKPD-FSCIL achieves substantial accuracy gains and zero-cost inference with only a minor and acceptable increase in the one-time, offline training cost per session.
\section{Conclusion}
In this paper, we identified a core limitation in prevailing FSCIL paradigms: by treating pretrained models as black boxes, they are forced into an inefficient trade-off between sacrificing plasticity and incurring unsustainable growth in parameters and inference cost.
We addressed this by proposing CKPD-FSCIL, a unified framework that systematically opens the black box to repurpose a model’s internal capacity. It integrates two synergistic, continuously adapting principles to manage knowledge at both the weight and layer levels: (1) Continuous Knowledge-Preserving Decomposition , which partitions weights into frozen and learnable subspaces via covariance-guided analysis; and (2) Continuous Adaptive Layer Selection, which offers a dynamic strategy for which layers to adapt across the network.
Extensive experiments across miniImageNet, CUB-200, and CIFAR-100 with diverse backbones, as well as large-scale ImageNet-1K, confirm that our unified approach achieves state-of-the-art accuracy, strong scalability, and exceptional efficiency. More broadly, our findings establish the reuse-and-reallocate paradigm as a more effective and sustainable alternative to the conventional add-or-freeze strategy for continual learning.

\bibliographystyle{IEEEtran}
\bibliography{main}

\begin{thebibliography}{10}
\providecommand{\url}[1]{#1}
\csname url@samestyle\endcsname
\providecommand{\newblock}{\relax}
\providecommand{\bibinfo}[2]{#2}
\providecommand{\BIBentrySTDinterwordspacing}{\spaceskip=0pt\relax}
\providecommand{\BIBentryALTinterwordstretchfactor}{4}
\providecommand{\BIBentryALTinterwordspacing}{\spaceskip=\fontdimen2\font plus
\BIBentryALTinterwordstretchfactor\fontdimen3\font minus \fontdimen4\font\relax}
\providecommand{\BIBforeignlanguage}[2]{{%
\expandafter\ifx\csname l@#1\endcsname\relax
\typeout{** WARNING: IEEEtran.bst: No hyphenation pattern has been}%
\typeout{** loaded for the language `#1'. Using the pattern for}%
\typeout{** the default language instead.}%
\else
\language=\csname l@#1\endcsname
\fi
#2}}
\providecommand{\BIBdecl}{\relax}
\BIBdecl

\bibitem{tao2020few}
X.~Tao, X.~Hong, X.~Chang, S.~Dong, X.~Wei, and Y.~Gong, ``Few-shot class-incremental learning,'' in \emph{CVPR}, 2020.

\bibitem{wang2024jarvis}
Z.~Wang, S.~Cai, A.~Liu, Y.~Jin, J.~Hou, B.~Zhang, H.~Lin, Z.~He, Z.~Zheng, Y.~Yang \emph{et~al.}, ``Jarvis-1: Open-world multi-task agents with memory-augmented multimodal language models,'' \emph{TPAMI}.

\bibitem{lioptimus}
Z.~Li, Y.~Xie, R.~Shao, G.~Chen, D.~Jiang, and L.~Nie, ``Optimus-1: Hybrid multimodal memory empowered agents excel in long-horizon tasks,'' in \emph{NeurIPS}, 2024.

\bibitem{zheng2025lifelong}
J.~Zheng, C.~Shi, X.~Cai, Q.~Li, D.~Zhang, C.~Li, D.~Yu, and Q.~Ma, ``Lifelong learning of large language model based agents: A roadmap,'' \emph{arXiv:2501.07278}, 2025.

\bibitem{li2024vision}
Z.~Li, Y.~Lv, Z.~Tu, D.~Shang, and H.~Qiao, ``Vision-language navigation with continual learning,'' \emph{arXiv:2409.02561}, 2024.

\bibitem{mccloskey1989catastrophic}
M.~McCloskey and N.~J. Cohen, \emph{Catastrophic interference in connectionist networks: The sequential learning problem}, ser. Psychology of Learning and Motivation.\hskip 1em plus 0.5em minus 0.4em\relax Elsevier, 1989, vol.~24.

\bibitem{goodfellow2013empirical}
I.~J. Goodfellow, M.~Mirza, D.~Xiao, A.~Courville, and Y.~Bengio, ``An empirical investigation of catastrophic forgetting in gradient-based neural networks,'' \emph{arXiv:1312.6211}, 2013.

\bibitem{snell2017prototypical}
J.~Snell, K.~Swersky, and R.~Zemel, ``Prototypical networks for few-shot learning,'' in \emph{NeurIPS}, 2017.

\bibitem{sung2018learning}
F.~Sung, Y.~Yang, L.~Zhang, T.~Xiang, P.~H.~S. Torr, and T.~M. Hospedales, ``Learning to compare: Relation network for few-shot learning,'' in \emph{CVPR}, 2018.

\bibitem{mermillod2013stability}
M.~Mermillod, A.~Bugaiska, and P.~Bonin, ``The stability-plasticity dilemma: Investigating the continuum from catastrophic forgetting to age-limited learning effects,'' 2013.

\bibitem{rebuffi2017icarl}
S.-A. Rebuffi, A.~Kolesnikov, G.~Sperl, and C.~H. Lampert, ``icarl: Incremental classifier and representation learning,'' in \emph{CVPR}, 2017.

\bibitem{hou2019learning}
S.~Hou, X.~Pan, C.~C. Loy, Z.~Wang, and D.~Lin, ``Learning a unified classifier incrementally via rebalancing,'' in \emph{CVPR}, 2019.

\bibitem{kirkpatrick2017overcoming}
J.~Kirkpatrick, R.~Pascanu, N.~Rabinowitz, J.~Veness, G.~Desjardins, A.~A. Rusu, K.~Milan, J.~Quan, T.~Ramalho, A.~Grabska-Barwinska \emph{et~al.}, ``Overcoming catastrophic forgetting in neural networks,'' \emph{National Academy of Sciences}, 2017.

\bibitem{shin2017continual}
H.~Shin, J.~K. Lee, J.~Kim, and J.~Kim, ``Continual learning with deep generative replay,'' 2017.

\bibitem{zhang2021few}
C.~Zhang, N.~Song, G.~Lin, Y.~Zheng, P.~Pan, and Y.~Xu, ``Few-shot incremental learning with continually evolved classifiers,'' in \emph{CVPR}, 2021.

\bibitem{yang2023neural}
Y.~Yang, H.~Yuan, X.~Li, Z.~Lin, P.~Torr, and D.~Tao, ``Neural collapse inspired feature-classifier alignment for few-shot class-incremental learning,'' in \emph{ICLR}, 2023.

\bibitem{li2024mamba}
X.~Li, Y.~Yang, J.~Wu, B.~Ghanem, L.~Nie, and M.~Zhang, ``Mamba-fscil: Dynamic adaptation with selective state space model for few-shot class-incremental learning,'' \emph{arXiv:2407.06136}, 2024.

\bibitem{park2024pre}
K.-H. Park, K.~Song, and G.-M. Park, ``Pre-trained vision and language transformers are few-shot incremental learners,'' in \emph{CVPR}, 2024.

\bibitem{ran2024brain}
H.~Ran, X.~Gao, L.~Li, W.~Li, S.~Tian, G.~Wang, H.~Shi, and X.~Ning, ``Brain-inspired fast-and slow-update prompt tuning for few-shot class-incremental learning,'' \emph{TNNLS}, 2024.

\bibitem{goswami2024calibrating}
D.~Goswami, B.~Twardowski, and J.~Van De~Weijer, ``Calibrating higher-order statistics for few-shot class-incremental learning with pre-trained vision transformers,'' in \emph{CVPR}, 2024.

\bibitem{d2023multimodal}
M.~D'Alessandro, A.~Alonso, E.~Calabr{\'e}s, and M.~Galar, ``Multimodal parameter-efficient few-shot class incremental learning,'' in \emph{ICCV}, 2023.

\bibitem{wang2024knowledge}
Y.~Wang, Y.~Wang, G.~Zhao, and X.~Qian, ``Knowledge adaptation network for few-shot class-incremental learning,'' \emph{arXiv:2409.11770}, 2024.

\bibitem{liu2024few}
C.~Liu, Z.~Wang, T.~Xiong, R.~Chen, Y.~Wu, J.~Guo, and H.~Huang, ``Few-shot class incremental learning with attention-aware self-adaptive prompt,'' \emph{arXiv:2403.09857}, 2024.

\bibitem{tian2024pl}
S.~Tian, L.~Li, W.~Li, H.~Ran, L.~Li, and X.~Ning, ``Pl-fscil: Harnessing the power of prompts for few-shot class-incremental learning,'' \emph{arXiv:2401.14807}, 2024.

\bibitem{li2017learning}
Z.~Li and D.~Hoiem, ``Learning without forgetting,'' \emph{TPAMI}, 2017.

\bibitem{zhang2023few}
J.~Zhang, L.~Liu, O.~Silven, M.~Pietik{\"a}inen, and D.~Hu, ``Few-shot class-incremental learning: A survey,'' \emph{arXiv:2308.06764}, 2023.

\bibitem{tian2024survey}
S.~Tian, L.~Li, W.~Li, H.~Ran, X.~Ning, and P.~Tiwari, ``A survey on few-shot class-incremental learning,'' \emph{Neural Networks}, 2024.

\bibitem{vinyals2016matching}
O.~Vinyals, C.~Blundell, T.~Lillicrap, and D.~Wierstra, ``Matching networks for one shot learning,'' in \emph{NeurIPS}, 2016.

\bibitem{ravi2017optimization}
S.~Ravi and H.~Larochelle, ``Optimization as a model for few-shot learning,'' in \emph{ICLR}, 2017.

\bibitem{yang2024corda}
Y.~Yang, X.~Li, Z.~Zhou, S.~L. Song, J.~Wu, L.~Nie, and B.~Ghanem, ``Corda: Context-oriented decomposition adaptation of large language models,'' \emph{arXiv:2406.05223}, 2024.

\bibitem{liu2022few}
H.~Liu, L.~Gu, Z.~Chi, Y.~Wang, Y.~Yu, J.~Chen, and J.~Tang, ``Few-shot class-incremental learning via entropy-regularized data-free replay,'' in \emph{ECCV}, 2022.

\bibitem{agarwal2022semantics}
A.~Agarwal, B.~Banerjee, F.~Cuzzolin, and S.~Chaudhuri, ``Semantics-driven generative replay for few-shot class incremental learning,'' in \emph{ACM MM}, 2022.

\bibitem{peng2022few}
C.~Peng, K.~Zhao, T.~Wang, M.~Li, and B.~C. Lovell, ``Few-shot class-incremental learning from an open-set perspective,'' in \emph{ECCV}, 2022.

\bibitem{yoon2020xtarnet}
S.~W. Yoon, D.-Y. Kim, J.~Seo, and J.~Moon, ``Xtarnet: Learning to extract task-adaptive representation for incremental few-shot learning,'' in \emph{ICML}, 2020.

\bibitem{chi2022metafscil}
Z.~Chi, L.~Gu, H.~Liu, Y.~Wang, Y.~Yu, and J.~Tang, ``Metafscil: A meta-learning approach for few-shot class incremental learning,'' in \emph{CVPR}, 2022.

\bibitem{zhou2022few}
D.-W. Zhou, H.-J. Ye, L.~Ma, D.~Xie, S.~Pu, and D.-C. Zhan, ``Few-shot class-incremental learning by sampling multi-phase tasks,'' \emph{TPAMI}, 2022.

\bibitem{cheraghian2021synthesized}
A.~Cheraghian, S.~Rahman, S.~Ramasinghe, P.~Fang, C.~Simon, L.~Petersson, and M.~Harandi, ``Synthesized feature based few-shot class-incremental learning on a mixture of subspaces,'' in \emph{ICCV}, 2021.

\bibitem{zhou2022forward}
D.-W. Zhou, F.-Y. Wang, H.-J. Ye, L.~Ma, S.~Pu, and D.-C. Zhan, ``Forward compatible few-shot class-incremental learning,'' in \emph{CVPR}, 2022.

\bibitem{tao2020topology}
X.~Tao, X.~Chang, X.~Hong, X.~Wei, and Y.~Gong, ``Topology-preserving class-incremental learning,'' in \emph{ECCV}, 2020.

\bibitem{joseph2022energy}
K.~Joseph, S.~Khan, F.~S. Khan, R.~M. Anwer, and V.~N. Balasubramanian, ``Energy-based latent aligner for incremental learning,'' in \emph{CVPR}, 2022.

\bibitem{lu2022geometer}
B.~Lu, X.~Gan, L.~Yang, W.~Zhang, L.~Fu, and X.~Wang, ``Geometer: Graph few-shot class-incremental learning via prototype representation,'' in \emph{ACM MM}, 2022.

\bibitem{chen2021incremental}
K.~Chen and C.-G. Lee, ``Incremental few-shot learning via vector quantization in deep embedded space,'' in \emph{ICLR}, 2021.

\bibitem{akyurek2021subspace}
A.~F. Aky{\"u}rek, E.~Aky{\"u}rek, D.~Wijaya, and J.~Andreas, ``Subspace regularizers for few-shot class incremental learning,'' in \emph{ICLR}, 2022.

\bibitem{song2023learning}
Z.~Song, Y.~Zhao, Y.~Shi, P.~Peng, L.~Yuan, and Y.~Tian, ``Learning with fantasy: Semantic-aware virtual contrastive constraint for few-shot class-incremental learning,'' in \emph{CVPR}, 2023.

\bibitem{ahmed2024orco}
N.~Ahmed, A.~Kukleva, and B.~Schiele, ``Orco: Towards better generalization via orthogonality and contrast for few-shot class-incremental learning,'' in \emph{CVPR}, 2024.

\bibitem{zou2024compositional}
Y.~Zou, S.~Zhang, Y.~Li, R.~Li \emph{et~al.}, ``Compositional few-shot class-incremental learning,'' in \emph{ICML}, 2024.

\bibitem{zhou2024delve}
H.~Zhou, Y.~Zou, R.~Li, Y.~Li, and K.~Xiao, ``Delve into base-novel confusion: redundancy exploration for few-shot class-incremental learning,'' in \emph{IJCAI}, 2024.

\bibitem{oh2024closer}
J.~Oh, S.~Baik, and K.~M. Lee, ``Closer: Towards better representation learning for few-shot class-incremental learning,'' in \emph{ECCV}, 2024, pp. 18--35.

\bibitem{hersche2022constrained}
M.~Hersche, G.~Karunaratne, G.~Cherubini, L.~Benini, A.~Sebastian, and A.~Rahimi, ``Constrained few-shot class-incremental learning,'' in \emph{CVPR}, 2022.

\bibitem{yang2023neural_arxiv}
Y.~Yang, H.~Yuan, X.~Li, J.~Wu, L.~Zhang, Z.~Lin, P.~Torr, D.~Tao, and B.~Ghanem, ``Neural collapse terminus: A unified solution for class incremental learning and its variants,'' \emph{arXiv:2308.01746}, 2023.

\bibitem{zhu2021self}
K.~Zhu, Y.~Cao, W.~Zhai, J.~Cheng, and Z.-J. Zha, ``Self-promoted prototype refinement for few-shot class-incremental learning,'' in \emph{CVPR}, 2021.

\bibitem{wang2024few}
Q.-W. Wang, D.-W. Zhou, Y.-K. Zhang, D.-C. Zhan, and H.-J. Ye, ``Few-shot class-incremental learning via training-free prototype calibration,'' in \emph{NeurIPS}.\hskip 1em plus 0.5em minus 0.4em\relax MIT Press, 2024.

\bibitem{dong2021few}
S.~Dong, X.~Hong, X.~Tao, X.~Chang, X.~Wei, and Y.~Gong, ``Few-shot class-incremental learning via relation knowledge distillation,'' in \emph{AAAI}, 2021.

\bibitem{cheraghian2021semantic}
A.~Cheraghian, S.~Rahman, P.~Fang, S.~K. Roy, L.~Petersson, and M.~Harandi, ``Semantic-aware knowledge distillation for few-shot class-incremental learning,'' in \emph{CVPR}, 2021.

\bibitem{zhao2023few}
L.~Zhao, J.~Lu, Y.~Xu, Z.~Cheng, D.~Guo, Y.~Niu, and X.~Fang, ``Few-shot class-incremental learning via class-aware bilateral distillation,'' in \emph{CVPR}, 2023.

\bibitem{lester2021power}
B.~Lester, R.~Al-Rfou, and N.~Constant, ``The power of scale for parameter-efficient prompt tuning,'' in \emph{EMNLP}, 2021.

\bibitem{li2021prefix}
X.~L. Li and P.~Liang, ``Prefix-tuning: Optimizing continuous prompts for generation,'' in \emph{IJCNLP}, 2021.

\bibitem{jia2022visual}
M.~Jia, L.~Tang, B.-C. Chen, C.~Cardie, S.~Belongie, B.~Hariharan, and S.-N. Lim, ``Visual prompt tuning,'' in \emph{ECCV}, 2022.

\bibitem{wang2022dualprompt}
Z.~Wang, Z.~Zhang, S.~Ebrahimi, R.~Sun, H.~Zhang, C.-Y. Lee, X.~Ren, G.~Su, V.~Perot, J.~Dy \emph{et~al.}, ``Dualprompt: Complementary prompting for rehearsal-free continual learning,'' in \emph{ECCV}, 2022.

\bibitem{houlsby2019parameter}
N.~Houlsby, A.~Giurgiu, S.~Jastrzebski, B.~Morrone, Q.~De~Laroussilhe, A.~Gesmundo, M.~Attariyan, and S.~Gelly, ``Parameter-efficient transfer learning for nlp,'' in \emph{ICML}, 2019.

\bibitem{he2022towards}
J.~He, C.~Zhou, X.~Ma, T.~Berg-Kirkpatrick, and G.~Neubig, ``Towards a unified view of parameter-efficient transfer learning,'' in \emph{ICLR}, 2022.

\bibitem{lei2023conditional}
T.~Lei, J.~Bai, S.~Brahma, J.~Ainslie, K.~Lee, Y.~Zhou, N.~Du, V.~Y. Zhao, Y.~Wu, B.~Li \emph{et~al.}, ``Conditional adapters: Parameter-efficient transfer learning with fast inference,'' in \emph{NeurIPS}, 2023.

\bibitem{hu2022lora}
E.~J. Hu, Y.~Shen, P.~Wallis, Z.~Allen-Zhu, Y.~Li, S.~Wang, L.~Wang, and W.~Chen, ``Lo{RA}: Low-rank adaptation of large language models,'' in \emph{ICLR}, 2022.

\bibitem{meng2024pissa}
F.~Meng, Z.~Wang, and M.~Zhang, ``Pissa: Principal singular values and singular vectors adaptation of large language models,'' \emph{arXiv:2404.02948}, 2024.

\bibitem{yuan2023asvd}
Z.~Yuan, Y.~Shang, Y.~Song, Q.~Wu, Y.~Yan, and G.~Sun, ``Asvd: Activation-aware singular value decomposition for compressing large language models,'' \emph{arXiv:2312.05821}, 2023.

\bibitem{liu2024dora}
S.-Y. Liu, C.-Y. Wang, H.~Yin, P.~Molchanov, Y.-C.~F. Wang, K.-T. Cheng, and M.-H. Chen, ``Dora: Weight-decomposed low-rank adaptation,'' in \emph{ICML}, 2024.

\bibitem{fisher1936use}
R.~A. Fisher, ``The use of multiple measurements in taxonomic problems,'' \emph{Annals of eugenics}, 1936.

\bibitem{mclachlan2005discriminant}
G.~J. McLachlan, \emph{Discriminant analysis and statistical pattern recognition}.\hskip 1em plus 0.5em minus 0.4em\relax John Wiley \& Sons, 2005.

\bibitem{he2023law}
H.~He and W.~J. Su, ``A law of data separation in deep learning,'' \emph{National Academy of Sciences}, 2023.

\bibitem{russakovsky2015imagenet}
O.~Russakovsky, J.~Deng, H.~Su, J.~Krause, S.~Satheesh, S.~Ma, Z.~Huang, A.~Karpathy, A.~Khosla, M.~Bernstein \emph{et~al.}, ``Imagenet large scale visual recognition challenge,'' \emph{IJCV}, 2015.

\bibitem{wah2011caltech}
C.~Wah, S.~Branson, P.~Welinder, P.~Perona, and S.~Belongie, ``The caltech-ucsd birds-200-2011 dataset,'' \emph{California Institute of Technology}, 2011.

\bibitem{krizhevsky2009learning}
A.~Krizhevsky, ``Learning multiple layers of features from tiny images,'' \emph{Citeseer}, 2009.

\bibitem{zhang2018mixup}
H.~Zhang, M.~Cisse, Y.~Dauphin, and D.~Lopez-Paz, ``mixup: Beyond empirical risk management,'' in \emph{ICLR}, 2018.

\bibitem{devries2017improved}
T.~DeVries, ``Improved regularization of convolutional neural networks with cutout,'' \emph{arXiv:1708.04552}, 2017.

\bibitem{dosovitskiy2020image}
A.~Dosovitskiy, L.~Beyer, A.~Kolesnikov, D.~Weissenborn, X.~Zhai, T.~Unterthiner, M.~Dehghani, M.~Minderer, G.~Heigold, S.~Gelly \emph{et~al.}, ``An image is worth 16x16 words: Transformers for image recognition at scale,'' in \emph{ICLR}, 2020.

\bibitem{gu2023mamba}
A.~Gu and T.~Dao, ``Mamba: Linear-time sequence modeling with selective state spaces,'' \emph{arXiv:2312.00752}, 2023.

\bibitem{radford2021learning}
A.~Radford, J.~W. Kim, C.~Hallacy, A.~Ramesh, G.~Goh, S.~Agarwal, G.~Sastry, A.~Askell, P.~Mishkin, J.~Clark \emph{et~al.}, ``Learning transferable visual models from natural language supervision,'' in \emph{ICML}, 2021.

\bibitem{wang2023improved}
Y.~Wang, G.~Zhao, and X.~Qian, ``Improved continually evolved classifiers for few-shot class-incremental learning,'' \emph{TCSVT}, 2023.

\bibitem{liu2021swin}
Z.~Liu, Y.~Lin, Y.~Cao, H.~Hu, Y.~Wei, Z.~Zhang, S.~Lin, and B.~Guo, ``Swin transformer: Hierarchical vision transformer using shifted windows,'' in \emph{ICCV}, 2021.

\bibitem{ahmad2022few}
T.~Ahmad, A.~R. Dhamija, S.~Cruz, R.~Rabinowitz, C.~Li, M.~Jafarzadeh, and T.~E. Boult, ``Few-shot class incremental learning leveraging self-supervised features,'' in \emph{CVPR}, 2022.

\bibitem{yang2022dynamic}
B.~Yang, M.~Lin, Y.~Zhang, B.~Liu, X.~Liang, R.~Ji, and Q.~Ye, ``Dynamic support network for few-shot class incremental learning,'' \emph{TPAMI}, 2022.

\bibitem{zou2022margin}
Y.~Zou, S.~Zhang, Y.~Li, and R.~Li, ``Margin-based few-shot class-incremental learning with class-level overfitting mitigation,'' in \emph{NeurIPS}, 2022.

\bibitem{deng2009imagenet}
J.~Deng, W.~Dong, R.~Socher, L.-J. Li, K.~Li, and L.~Fei-Fei, ``Imagenet: A large-scale hierarchical image database,'' in \emph{CVPR}, 2009.

\bibitem{smith2023coda}
J.~S. Smith, L.~Karlinsky, V.~Gutta, P.~Cascante-Bonilla, D.~Kim, A.~Arbelle, R.~Panda, R.~Feris, and Z.~Kira, ``Coda-prompt: Continual decomposed attention-based prompting for rehearsal-free continual learning,'' in \emph{CVPR}, 2023.

\bibitem{zhou2024expandable}
D.-W. Zhou, H.-L. Sun, H.-J. Ye, and D.-C. Zhan, ``Expandable subspace ensemble for pre-trained model-based class-incremental learning,'' in \emph{CVPR}, 2024.

\bibitem{wu2019large}
Y.~Wu, Y.~Chen, L.~Wang, Y.~Ye, Z.~Liu, Y.~Guo, and Y.~Fu, ``Large scale incremental learning,'' in \emph{CVPR}, 2019.

\bibitem{kukleva2021generalized}
A.~Kukleva, H.~Kuehne, and B.~Schiele, ``Generalized and incremental few-shot learning by explicit learning and calibration without forgetting,'' in \emph{ICCV}, 2021.

\end{thebibliography}

\end{document}